# COUNTERFACTUAL EXPLANATIONS WITHOUT OPENING THE BLACK BOX: AUTOMATED DECISIONS AND THE GDPR

Sandra Wachter,[*] Brent Mittelstadt,[**] & Chris Russell[***]

---

[*] Oxford Internet Institute, University of Oxford, 1 St. Giles, Oxford, OX1 3JS, UK and The Alan Turing Institute, British Library, 96 Euston Road, London, NW1 2DB, UK. E-mail: sandra.wachter@oii.ox.ac.uk. This work was supported by The Alan Turing Institute under the EPSRC grant EP/N510129/1.

[**] Oxford Internet Institute, University of Oxford, 1 St. Giles, Oxford, OX1 3JS, UK, The Alan Turing Institute, British Library, 96 Euston Road, London, NW1 2DB, UK, Department of Science and Technology Studies, University College London, 22 Gordon Square, London, WC1E 6BT, UK.

[***] The Alan Turing Institute, British Library, 96 Euston Road, London, NW1 2DB, UK, Department of Electrical and Electronic Engineering, University of Surrey, Guildford, GU2 7HX, UK.

## TABLE OF CONTENTS

## I. INTRODUCTION

There has been much discussion of the existence of a "right to explanation" in the EU General Data Protection Regulation ("GDPR"), and its merits and disadvantages.[1] Attempts to implement a right to explanation that opens the "black box" to provide insight into the internal decision-making process of algorithms face four major legal and technical barriers. First, a legally binding right to explanation does not exist in the GDPR.[2] Second, even if legally binding, the right would only apply in limited cases (when a negative decision was solely automated and had legal or other similar significant effects).[3] Third, explaining the functionality of complex algorithmic decision-making systems and their rationale in specific cases is a technically challenging problem.[4] Explanations may likewise offer little meaningful information to data subjects, raising questions about their value.[5] Finally, data controllers have an interest in not sharing details of their algorithms to avoid disclosing trade secrets, violating the rights and freedoms of others (e.g. privacy), and allowing data subjects to game or manipulate the decision-making system.[6]

---

[1] *See, e.g.*, Sandra Wachter, Brent Mittelstadt & Luciano Floridi, *Why a Right to Explanation of Automated Decision-Making Does Not Exist in the General Data Protection Regulation*, 7 INT'L DATA PRIV. LAW 76, 79–90 (2017); Isak Mendoza & Lee A. Bygrave, *The Right Not to Be Subject to Automated Decisions Based on Profiling*, in EU INTERNET LAW: REGULATION AND ENFORCEMENT (Tatiani Synodinou et al. eds., 2017), https://papers.ssrn.com/abstract=2964855 [https://perma.cc/XV3T-G98W]; Lilian Edwards & Michael Veale, *Slave to the Algorithm? Why a 'Right to Explanation' is Probably Not the Remedy You are Looking For*, 16 DUKE L. TECH. REV. 18, 18–19 (2017); Tae Wan Kim & Bryan Routledge, *Algorithmic Transparency, a Right to Explanation, and Placing Trust*, SQUARESPACE (June 2017), https://static1.squarespace.com/static/592ee286d482e908d35b8494/t/59552415579fb3 0c014cd06c/1498752022120/Algorithmic+transparency%2C+a+right+to+explanation+ and+trust+%28TWK%26BR%29.pdf [https://perma.cc/K53W-GVN2]; Gianclaudio Malgieri & Giovanni Comandé, *Why a Right to Legibility of Automated Decision-Making Exists in the General Data Protection Regulation*, 7 INT'L DATA PRIV. L. 243, 246–47 (2017); Bryce Goodman & Seth Flaxman, *EU Regulations on Algorithmic Decision-Making and a "Right to Explanation,"* ARXIV:1606.08813, at 6–7 (2016), http://arxiv.org/abs/1606.08813 [https://perma.cc/5ZTR-WG8R]; Andrew Selbst & Julia Powles, *Meaningful Information and the Right to Explanation*, 7 INT'L DATA PRIV. L. 233, 233–34 (2017).

[2] Wachter, Mittelstadt & Floridi, *supra* note 1, at 79; Kim & Routledge, *supra* note 1, at 3.

[3] Wachter, Mittelstadt & Floridi, *supra* note 1, at 78.

[4] *See, e.g.*, Wachter, Mittelstadt & Floridi, *supra* note 1, at 77; Edwards & Veale, *supra* note 1, at 22; Joshua A. Kroll et al., *Accountable Algorithms*, 165 U. PA. L. REV. 633, 638 (2016); Tal Zarsky, *Transparent Predictions*, 2013 U. ILL. L. REV. 1503, 1519–20 (2013).

[5] Jenna Burrell, *How the Machine "Thinks:" Understanding Opacity in Machine Learning Algorithms*, BIG DATA & SOC., Jan.–June 2016, at 5; Kroll et al., *supra* note 4, at 638.

Despite these difficulties, the social and ethical value (and perhaps responsibility) of offering explanations to affected data subjects remains unaffected. One significant point has been neglected in this discussion. An explanation of automated decisions, both as envisioned by the GDPR and in general, does not necessarily hinge on the general public understanding of how algorithmic systems function. Even though such interpretability is of great importance and should be pursued, explanations can, in principle, be offered without opening the "black box." Looking at explanations as a means to help a data subject *act* rather than merely understand, one could gauge the scope and content of explanations according to the specific goal or action they are intended to support.

Explanations can serve many purposes. To investigate the potential scope of explanations, it seems reasonable to start from the perspective of the data subject, which is the natural person whose data is being collected and evaluated. We propose three aims for explanations to assist data subjects: (1) to inform and help the subject understand why a particular decision was reached, (2) to provide grounds to contest adverse decisions, and (3) to understand what could be changed to receive a desired result in the future, based on the current decision-making model. As we show, the GDPR offers little support to achieve any of these aims. However, none hinge on explaining the internal logic of automated decision-making systems.

Building trust is essential to increase societal acceptance of algorithmic decision-making. As a solution to close current gaps in transparency and accountability that undermine trust between data controllers and data subjects,[7] we propose to move beyond the limitations of the GDPR. We argue that counterfactuals should be used as a means to provide explanations for individual decisions.

---

[6] Burrell, *supra* note 5, at 3; Brenda Reddix-Smalls, *Credit Scoring and Trade Secrecy: An Algorithmic Quagmire or How the Lack of Transparency in Complex Financial Models Scuttled the Finance Market*, 12 U.C. DAVIS BUS. L.J. 87, 94 (2011); Mike Ananny & Kate Crawford, *Seeing without knowing: Limitations of the Transparency Ideal and its Application to Algorithmic Accountability*, NEW MEDIA & SOC., 2016, at 8, http://journals.sagepub.com/doi/full/10.1177/1461444816676645
[https://perma.cc/3HF6-G9DS]; Roger A. Ford & W. Nicholson Price II, *Privacy and Accountability in Black-Box Medicine*, 23 MICH. TELECOMM. TECH. REV. 1, 3 (2016); Frank A. Pasquale, *Restoring Transparency to Automated Authority*, 9 J. TELECOMM. HIGH TECH. L. 235, 237 (2011).
[7] Wachter, Mittelstadt & Floridi, *supra* note 1, at 78; Mendoza & Bygrave, *supra* note 1, at 97.

*Unconditional counterfactual explanations* should be given for positive and negative automated decisions, regardless of whether the decisions are solely (as opposed to predominantly) automated or produce legal or other significant effects. This approach provides data subjects with meaningful explanations to understand a given decision, grounds to contest it, and advice on how the data subject can change his or her behaviour or situation to possibly receive a desired decision (e.g. loan approval) in the future without facing the severely limited applicability imposed by the GDPR's definition of automated individual decision-making.[8]

In this paper, we present the concept of unconditional counterfactual explanations as a novel type of explanation of automated decisions that overcomes many challenges facing current work on algorithmic interpretability and accountability. We situate counterfactuals in the philosophical history of knowledge, as well as historical and modern research on interpretability and fairness in machine learning. Based on the potential advantages offered to data subjects by counterfactual explanations, we then assess their alignment with the GDPR's numerous provisions concerning automated decision-making. Specifically, we examine whether the GDPR offers support for explanations that aim to help data subjects understand the scope of automated decision-making as well as the rationale of specific decisions, explanations to contest decisions, and explanations that offer guidance on how data subjects can change their behaviour to receive a desired result. We conclude that unconditional counterfactual explanations can bridge the gap between the interests of data subjects and data controllers that otherwise acts as a barrier to a legally binding right to explanation.

## II. COUNTERFACTUALS

Counterfactual explanations take a similar form to the statement:

"You were denied a loan because your annual income was £30,000. If your income had been £45,000, you would have been offered a loan."

---

[8] Wachter, Mittelstadt & Floridi, *supra* note 1, at 87–88; Mendoza & Bygrave, *supra* note 1, at 83; Edwards & Veale, *supra* note 1, at 22.

Here the statement of decision is followed by a counterfactual, or statement of how the world would have to be different for a desirable outcome to occur. Multiple counterfactuals are possible, as multiple desirable outcomes can exist, and there may be several ways to achieve any of these outcomes. The concept of the "closest possible world," or the smallest change to the world that can be made to obtain a desirable outcome, is key throughout the discussion of counterfactuals. In many situations, providing several explanations covering a range of diverse counterfactuals corresponding to relevant or informative "close possible worlds" rather than "the closest possible world" may be more helpful. Knowing the smallest possible change to a variable or set of variables to arrive at a different outcome may not always be the most helpful type of counterfactual. Rather, relevance will depend also upon other case-specific factors, such as the mutability of a variable or real world probability of a change.[9]

In the existing literature, "explanation" typically refers to an attempt to convey the internal state or logic of an algorithm that leads to a decision.[10] In contrast, counterfactuals describe a dependency on the external facts that led to that decision. This is a crucial distinction. In modern machine learning, the internal state of the algorithm can consist of millions of variables intricately connected in a large web of dependent behaviours.[11] Conveying this state to a layperson in a way that allows them to reason about the behaviour of an algorithm is extremely challenging.[12]

The machine learning and legal communities have both taken relatively restricted views on what passes for an explanation. The machine learning community has been primarily concerned with debugging[13] and conveying approximations of algorithms that programmers or researchers could use to understand which features are important[14] while law and ethics scholars have been more concerned with understanding the internal logic of decisions as a means to assess their lawfulness (e.g. prevent discriminatory outcomes), contest them, increase accountability generally, and clarify liability.[15]

---

[9] *See infra*, Section II.A.

[10] *See* Burrell, *supra* note 5, at 1.

[11] *See, e.g.*, Kaiming He et al., *Deep Residual Learning for Image Recognition*, *in* PROCEEDINGS OF THE IEEE CONFERENCE ON COMPUTER VISION AND PATTERN RECOGNITION 770–78 (2016).

[12] *See* Burrell, *supra* note 5, at 1; Zachary C. Lipton, *The Mythos of Model Interpretability*, *in* 2016 WORKSHOP ON HUMAN INTERPRETABILITY IN MACHINE LEARNING 96, http://zacklipton.com/media/papers/mythos_model_interpretability_lipton2016.pdf [https://perma.cc/4JVZ-7T6D].

[13] Osbert Bastani, Carolyn Kim & Hamsa Bastani, *Interpretability via Model Extraction*, ARXIV:1706.09773, at 1 (2017), https://arxiv.org/pdf/1611.07450.pdf [https://perma.cc/8J3J-RE2T].

As such, the proposal made here for counterfactuals as explanations lies outside of the taxonomies of explanations proposed previously in machine learning, legal, and ethical literature. In contrast, as we discuss in the next section, analytic philosophy has taken a much broader view of knowledge and how counterfactuals can be used as justifications of beliefs.[16]

## A. HISTORIC CONTEXT AND THE PROBLEM OF KNOWLEDGE

Analytic Philosophy has a long history of analysing the necessary conditions for propositional knowledge.[17] Expressions of the type "*S*

---

[14] Marco Tulio Ribeiro, Sameer Singh & Carlos Guestrin, *Why Should I Trust You?: Explaining the Predictions of Any Classifier*, *in* PROCEEDINGS OF THE 22ND ACM SIGKDD INTERNATIONAL CONFERENCE ON KNOWLEDGE DISCOVERY AND DATA MINING 1135 (2016); Ramprasaath R. Selvaraju et al., *Grad-CAM: Why Did You Say That?*, ARXIV:1611.07450, at 1 (2016) , https://arxiv.org/abs/1611.07450 [https://perma.cc/AA8F-45XJ]; Karen Simonyan, Andrea Vedaldi & Andrew Zisserman, *Deep inside convolutional networks: Visualising Image Classification Models and Saliency Maps*, ARXIV:1312.6034, at 1 (2013), https://arxiv.org/abs/1312.6034 [https://perma.cc/Y85R-X9UE].

[15] *See, e.g.*, Finale Doshi-Velez et al., *Accountability of AI Under the Law: The Role of Explanation*, ARXIV:1711.01134, at 1 (2017); Finale Doshi-Velez, Ryan Budish & Mason Kortz, *The Role of Explanation in Algorithmic Trust*, TRUSTWORTHY ALGORITHMIC DECISION-MAKING 2, http://trustworthy-algorithms.org/whitepapers/Finale%20Doshi-Velez.pdf [https://perma.cc/4L88-V58A]; Mireille Hildebrandt, *The Dawn of a Critical Transparency Right for the Profiling Era*, *in* DIGITAL ENLIGHTENMENT YEARBOOK 2012 41 (Jacques Bus et al. eds., 2012); Tim Miller, *Explanation in Artificial Intelligence: Insights from the Social Sciences*, ARXIV:1706.07269, at 3 (2017); Pasquale, *supra* note 6, at 236; Danielle Keats Citron & Frank A. Pasquale, *The Scored Society: Due Process for Automated Predictions*, 89 WASH. L. REV. 1, 6–7 (2014), https://papers.ssrn.com/abstract=2376209 [https://perma.cc/9CXY-DBTN]; Tal Zarsky, *Transparent Predictions*, 2013 U. ILL. L. REV. 1503, 1506–09 (2013), https://papers.ssrn.com/sol3/papers.cfm?abstract_id=2324240 [https://perma.cc/F8FC-YDJG]; Tal Zarsky, *The Trouble with Algorithmic Decisions: An Analytic Road Map to Examine Efficiency and Fairness in Automated and Opaque Decision Making*, 41 SCI. TECH. HUM. VALUES 118, 118–132 (2016).

[16] *See, e.g.*, DAVID LEWIS, COUNTERFACTUALS 1–4, 84–91 (1973); David Lewis, *Counterfactuals and Comparative Possibility*, 2 J. PHIL. LOGIC 418, 418–446 (1973); Peter Lipton, *Contrastive Explanation*, 27 ROYAL INST. PHIL. SUPP. 247, 247 (1990); ROBERT NOZICK, PHILOSOPHICAL EXPLANATIONS 172–74 (1981).

[17] *See generally* ALFRED JULES AYER, THE PROBLEM OF KNOWLEDGE (1956).

knows that *p*" constitute knowledge, where *S* refers to the knowing subject, and *p* to the proposition that is known. Traditional approaches, which conceive of knowledge as "justified true belief," conceive of three necessary conditions for knowledge: truth, belief, and justification.[18] According to this tripartite approach, in order to know something, it is not enough to simply believe that something is true: rather, you must also have a good reason for believing it.[19] The relevance of this approach comes from the observation that this form of justification of beliefs can serve as a type of explanation,[20] as it is fundamentally a reason that a belief is held and therefore serves as an answer to the question, "Why do you believe *X*?" Understanding the different forms these justifications can take opens the door to a broader class of explanations than previously encountered in interpretability research.

Although influential, "justified true belief" has faced much criticism[21] and inspired substantial analysis of modifications to this tripartite approach as well as proposals for additional necessary conditions for a proposition to constitute knowledge.[22] Modal conditions, including safety[23] and sensitivity,[24] have been proposed as necessary additions to the tripartite built on counterfactual relations.[25]

Sosa[26] as well as Ichikawa and Steup[27] define sensitivity as:

If *p* were false, *S* would not believe that *p*.

---

[18] *See generally* Edmund L. Gettier, *Is Justified True Belief Knowledge?*, 23 ANALYSIS 121 (1963); Julien Dutant, *The Legend of the Justified True Belief Analysis*, 29 PHIL. PERSP. 95 (2015).

[19] *See* Gettier, *supra* note 18, at 121.

[20] *See* NOZICK, *supra* note 16, at 174.

[21] *See, e.g.*, Dutant, *supra* note 18, at 95; Mark Kaplan, *It's Not What You Know that Counts*, 82 J. PHIL. 350, 350 (1985).

[22] Jonathan Ichikawa & Matthias Steup, *The Analysis of Knowledge*, *in* STANFORD ENCYCLOPEDIA OF PHILOSOPHY ARCHIVE (Fall 2017 ed.), https://plato.stanford.edu/archives/fall2017/entries/knowledge-analysis/ [https://perma.cc/6CXB-FTJV].

[23] *See* Ernest Sosa, *How to Defeat Opposition to Moore*, 13 PHIL. PERSP. 141, 141–43 (1999).

[24] *See* Jonathan Ichikawa, *Quantifiers, Knowledge, and Counterfactuals*, 82 PHIL. PHENOMENOLOGICAL RES. 287, 287 (2011); *see also* NOZICK, *supra* note 16, at 172–74.

[25] *See* Ichikawa & Steup, *supra* note 22, Section 5 (reviewing these concepts and their criticisms).

[26] Sosa, *supra* note 23, at 141.

[27] Ichikawa & Steup, *supra* note 22, Section 5.1.

Here, the statement "If $p$ were false" is a counterfactual defining a "possible world" close to the world in which $p$ is true.[28] The sensitivity condition suggests that "in the nearest possible worlds in which not-$p$, the subject does not believe that $p$."[29] Our notion of counterfactual explanations hinges upon the related concept:

> If $q$ were false, $S$ would not believe $p$.

We claim that in this case, $q$ serves as an explanation of $S$'s belief in $p$, inasmuch as $S$ only holds belief $p$ while $q$ is true, and that changing $q$ would also cause $S$'s belief to change. A key point is that such statements only describe $S$'s beliefs, which need not reflect reality.[30] As such, these statements can be made without knowledge of any causal relationship between $q$ and $p$.

We define Counterfactual Explanations as statements taking the form:

> Score $p$ was returned because variables $V$ had values ($v_1$, $v_2$,...) associated with them. If $V$ instead had values ($v_1'$, $v_2'$,...), and all other variables had remained constant, score $p'$ would have been returned.

While many such explanations are possible, an ideal counterfactual explanation would alter values as little as possible and represent a closest world under which score $p'$ is returned instead of $p$. The notion of a "closest possible world" is thus implicit in our definition.

We define our version of counterfactuals perhaps most resembles a structural equations approach in execution by identifying alterations to variables. This approach is more similar to Pearl's "mini-surgeries"[31] than Lewis' "miracles."[32] In any case, our approach does not rely on knowledge of the causal structure of the world,[33] or suggest which context-dependent metric of distance between worlds is preferable to establish causality.[34] In many situations, it will be more informative to provide a diverse set of counterfactual explanations, corresponding to different choices of nearby possible worlds for which the counterfactual

---

holds or a preferred outcome is delivered, rather than a theoretically ideal counterfactual describing the "closest possible world" according to a preferred distance metric.[35] Case-specific considerations will be relevant to the choice of distance metric and a "sufficient" and "relevant" set of counterfactual explanations. Such considerations may include the capabilities of the individual concerned, sensitivity, mutability of the variables involved in a decision, and ethical or legal requirements for disclosure.[36]

Similarly, counterfactuals that describe changes to multiple variables within the model can be provided. These would represent possible futures brought about by changes to the individual's circumstances. As an example, the impact of changes in income could be calculated in combination with changes to career, thereby ensuring the counterfactual represents a realistic possible world.

## B. EXPLANATIONS IN A.I. AND MACHINE LEARNING

Much of the early work in A.I. on explaining the decisions made by expert or rule-based systems focused on classes of explanation closely related to counterfactuals. For example, Gregor and Benbasat[37] offer the following example of what they call a type 1 explanation:

> Q: Why is a tax cut appropriate?
> A: Because a tax cut's preconditions are high inflation and trade deficits, and current conditions include these factors.

---

[35] The merits of different metrics of distance between possible worlds have long been debated in philosophy without the emergence of consensus. Meaningfully addressing this debate goes beyond the scope of this paper which proposes a method for counterfactual explanations, but will be explored in future work. For further discussion of distance metrics and counterfactuals, *see* LEWIS, *supra* note 16, at 8–15; Ernest W. Adams, *On the Rightness of Certain Counterfactuals*, 74 PAC. PHIL. Q. 1, 1–8 (1993); Kment, *supra* note 34, at 262.

[36] A discussion of appropriate metrics for making these choices goes beyond the scope of this paper, but will be addressed in future work. With that said, relevant philosophical discussion can be found on determining relevance of possible causal or contrastive explanations, counterfactuals, and distance metrics. *See, e.g.*, Peter Lipton, *Contrastive Explanation*, 27 ROYAL INST. PHIL. SUPP. 247, 254–65 (1990); Adams, *supra* note 35, at 1–8.

[37] Shirley Gregor & Izak Benbasat, *Explanations from Intelligent Systems: Theoretical Foundations and Implications for Practice*, 23 MIS Q. 497, 503 (1999).

Buchanan and Shortliffe[38] offer a similar example:

> **RULE009** IF:
> 1) The gram stain of the organism is gramneg, and
> 2) The morphology of the organism is coccus
> THEN: There is strongly suggestive evidence (.8) that the
> identity of the organism is Neisseria

As is typical in early A.I., questions we now recognise as hard such as "How do we decide inflation is high?" or "Why are these the preconditions of a tax cut?" are assumed to have been addressed by humans, and are not discussed as part of the explanation.[39] As such, the explanations do not provide insight into what people in machine learning think of as the internal logic of black box classifiers. In fact, the first example can be rewritten as two diverse counterfactual statements:

> "If inflation was lower, a tax cut would not be recommended."

> "If there was no trade deficit, a tax cut would not be recommended."

While the second example is closely related to the counterfactual:[40]

> "If the gram stain was negative or the morphology was not coccus, the algorithm would not be confident that the organism is Neisseria."

The most important difference between these approaches and counterfactuals is that counterfactuals continue functioning in an end-to-end integrated approach. If the gram stain and morphology in the MYCIN example were also determined by the algorithm, counterfactuals would automatically return a close sample with a different classification, while these early methods could not be applied to such involved scenarios.

---

[38] BRUCE G. BUCHANAN & EDWARD D. SHORTLIFFE, RULE-BASED EXPERT SYSTEMS: THE MYCIN EXPERIMENTS OF THE STANFORD HEURISTIC PROGRAMMING PROJECT 344 (1984).

[39] *See, e.g.*, Gregor & Benbasat, *supra* note 37, at 503.

[40] However, they are not logically equivalent. The example from MYCIN differs in that it is still possible that some samples that are either gram positive or have a different morphology could still be classified as Neisseria.

As focus has switched from A.I. and logic-based systems towards machine learning tasks such as image recognition, the notion of an explanation has come to refer to providing insight into the internal state of an algorithm, or to human-understandable approximations of the algorithm.[41] As such, the most related machine learning work to these, and to ours, is Martens and Provost.[42] Uniquely among other works in machine learning, it shares our interest in making interventions to alter the outcome of classifier responses. However, the work is firmly linked to the problem of document classification, and the only interventions it proposes involve the removal of words from documents to stop websites from being classified as "adult."[43] The heuristic proposed cannot be easily generalised to either continuous variables,[44] or even the addition of words to documents.

The majority of works in machine learning on explanations and interpreting models concern themselves with generating simple models as local approximations of decisions.[45] Generally, the idea is to create a simple human-understandable approximation of a decision-making algorithm that accurately models the decision given the current inputs, but may be arbitrarily bad for different inputs.[46] However, there are numerous difficulties with treating these approaches as explanations suitable for a lay data subject.

In general, it is unclear if these models are interpretable by non-experts. They make a three-way trade-off between the quality of the approximation, the ease of understanding the function, and the size of the domain for which the approximation is valid.[47] As we show in Appendix 1, these local models can produce widely varying estimates of the importance of variables even in simple scenarios such as the single

---

variable case, making it extremely difficult to reason about how a function varies as the inputs change. Moreover, the utility of such approaches outside of model debugging by expert programmers is unclear. Research has yet to be conducted on how to convey the various limitations and unreliabilities of these approaches to a lay audience in such a way that they can make use of such explanations.

In contrast, counterfactual explanations are intentionally restricted. They are crafted in such a way as to provide a minimal amount of information capable of altering a decision, and they do not require the data subject to understand any of the internal logic of a model in order to make use of it. The downside to this is that individual counterfactuals may be overly restrictive. A single counterfactual may show how a decision is based on certain data that is both correct and unable to be altered by the data subject before future decisions, even if other data exist that could be amended for a favourable outcome. This problem could be resolved by offering multiple diverse counterfactual explanations to the data subject.

## C. ADVERSARIAL PERTURBATIONS AND COUNTERFACTUAL EXPLANATIONS

The techniques used to generate counterfactual explanations on deep networks such as resnet[48] are already widely studied in the machine learning literature under the name of "Adversarial Perturbations."[49] In these works, algorithms capable of computing counterfactuals are used to confuse existing classifiers by generating a synthetic data point close to an existing one such that the new synthetic data point is classified differently than the original one.[50]

One strength of counterfactuals is that they can be efficiently and effectively computed by applying standard techniques, even to cutting-edge architectures. Some of the largest and deepest neural networks are used in the field of computer vision, particularly in image labelling tasks such as ImageNet.[51] These classifiers have been shown to be particularly vulnerable to a type of attack referred to as "Adversarial Perturbation" where small changes to a given image can result in the image being assigned to an entirely different class. For example, DeepFool[52] defines an adverse perturbation of an image $x$, given a classifier, as the smallest change to $x$ such that the classification changes. Essentially, this is a counterfactual by a different name. Finding a closest possible world to $x$ such that the classification changes is, under the right choice of distance function, the same as finding the smallest change to $x$.

---

[48] *See* He et al., *supra* note 11, at 770.

[49] *See* Ian J. Goodfellow, Jonathon Shlens & Christian Szegedy, *Explaining and Harnessing Adversarial Examples*, ARXIV:1412.6572 , at 1 (2014), https://arxiv.org/pdf/1412.6572.pdf [https://perma.cc/64BR-WVE7]; Seyed-Mohsen Moosavi-Dezfooli, Alhussein Fawzi & Pascal Frossard, *Deepfool: A Simple and Accurate Method to Fool Deep Neural Networks*, *in* PROCEEDINGS OF THE IEEE CONFERENCE ON COMPUTER VISION AND PATTERN RECOGNITION 2574–82 (2016); Christian Szegedy et al., *Intriguing Properties of Neural Networks*, ARXIV:1312.6199, at 2 (2013) , https://arxiv.org/pdf/1312.6199.pdf [https://perma.cc/K37R-6NP2].

[50] *See* Goodfellow, Shlens & Szegedy, *supra* note 49, at 1; Moosavi-Dezfooli, Fawzi & Frossard, *supra* note 49, at 2574–82; Szegedy et al., *supra* note 49, at 2.

Importantly, none of the standard works on Adversarial Perturbations make use of appropriate distance functions, and the majority of such approaches tend to favour making small changes to many variables, instead of providing sparse human interpretable solutions that modify only a few variables.[53] Despite this, efficient computation of counterfactuals and Adversarial Perturbations is made possible by virtue of state-of-the-art algorithms being differentiable. Many optimisation techniques proposed in the Adversarial Perturbation literature are directly applicable to this problem, making counterfactual generation efficient.

One of the more challenging aspects of Adversarial Perturbations is that these small perturbations of an image are barely human perceptible, but result in drastically different classifier responses.[54] Informally, this appears to happen because the newly generated images do not lie in the "space of real-images," but slightly outside it.[55] This phenomenon serves as an important reminder that when computing counterfactuals by searching for a close possible world, it is at least as important that the solution found comes from a "possible world" as it is that it is close to the starting example. Further research into how data from high-dimensional and highly-structured spaces, such as natural images, can be characterised is needed before counterfactuals can be reliably used as explanations in these spaces.

---

[51] *See* Jia Deng et al., *Imagenet: A Large-Scale Hierarchical Image Database*, *in* IEEE CONFERENCE ON COMPUTER VISION AND PATTERN RECOGNITION 248–55 (2009).

[52] *See* Moosavi-Dezfooli, Fawzi & Frossard, *supra* note 49, at 2574.

[53] *See* Jiawei Su, Danilo Vasconcellos Vargas & Sakurai Kouichi, *One Pixel Attack for Fooling Deep Neural Networks*, ARXIV:1710.08864, at 8–9 (2017), http://arxiv.org/abs/1710.08864 [https://perma.cc/5F2N-JBJF].

[54] Niki Kilbertus et al., *Avoiding Discrimination through Causal Reasoning*, ARXIV:1706,02744, at 1 (2017), https://arxiv.org/pdf/1706.02744.pdf [https://perma.cc/8GZQ-PRSJ].

[55] *See generally* Simant Dube, *High Dimensional Spaces, Deep Learning and Adversarial Examples*, ARXIV:1801.00634 (2018), https://arxiv.org/pdf/1801.00634.pdf [https://perma.cc/8FE5-RY4X] (presenting preliminary investigation of this matter).

### D. CAUSALITY AND FAIRNESS

Several works have approached the problem of guaranteeing that algorithms are fair, i.e. that they do not exhibit a bias towards particular ethnic, gender, or other protected groups, using causal reasoning[56] and counterfactuals.[57] Kusner et al.[58] consider counterfactuals where the subject belongs to a different race or sex, and require that the decision made remain the same under such a counterfactual for it to be considered fair. In contrast, we consider counterfactuals in which the decision differs from its current state.

Many works have suggested that transparency might be a useful tool for enforcing fairness. While it is unclear how counterfactuals could be used for this purpose, it is also unclear if any form of explanation of individual decisions can in fact help. Grgic-Hlaca et al. [59] showed how understandable models can easily mislead our intuitions, and that predominantly using features people believed to be fair slightly *increased* the racism exhibited by algorithms, while decreasing accuracy. In general, the best tools for uncovering systematic biases are likely to be based upon large-scale statistical analysis and not upon explanations of individual decisions.[60]

With that said, counterfactuals can provide evidence that an algorithmic decision is affected by a protected variable (e.g. race), and that it may therefore be discriminatory.[61] For the types of distance function we consider in the next section, if the counterfactuals found change someone's race, then the treatment of that individual is dependent on race. However, the converse statement is *not* true. Counterfactuals which do not modify a protected attribute cannot be used as evidence that the attribute was irrelevant to the decision. This is because counterfactuals describe only some of the dependencies between a particular decision and specific external facts. This can be seen clearly in Section III.A, where the counterfactuals proposed for a particular classifier involve 'black' people changing their race, while not suggesting that 'white' people's race should be varied.

---

[56] *See* Kilbertus et al., *supra* note 54, at 1.

[57] *See* Matt J. Kusner et al., *Counterfactual Fairness*, ArXiv:1703.06856, at 16 (2017), https://arxiv.org/pdf/1703.06856.pdf [https://perma.cc/4SVN-7J9D].

[58] *Id.*

[59] Nina Grgic-Hlaca et al., *The Case for Process Fairness in Learning: Feature Selection for Fair Decision Making*, *in* NIPS Symposium on Machine Learning and the Law 8 (2016).

[60] *See* Andrea Romei & Salvatore Ruggieri, *A Multidisciplinary Survey on Discrimination Analysis*, 29 Knowledge Engineering Rev. 582, 617 (2014).

[61] Establishing the influence of a protected variable on a decision does not, by itself, prove that illegal discrimination has occurred. Mitigating factors may exist which justify the usage of a protected attribute. *See, e.g.*, Solon Barocas & Andrew D. Selbst, *Big Data's Disparate Impact*, 104 Cal. L. Rev. 671, 676 (2016) (discussing disparate treatment in American anti-discrimination law).

## III. GENERATING COUNTERFACTUALS

In the following section, we give examples of how meaningful counterfactuals can be easily computed. Many of the standard classifiers of machine learning (including Neural Networks, Support Vector Machines, and Regressors) are trained by finding the optimal set of weights $w$ that minimises an objective over a set of training data.

$$\arg\min_w \ell(f_w(x_i), y_i) + \rho(w)$$

Equation 1

Where $y_i$ is the label for data point $x_i$ and $\rho(\cdot)$ is a regularizer over the weights. We wish to find a counterfactual $x'$ as close to the original point $x_i$ as possible such that $f_w(x')$ is equal to a new target $y'$. We can find $x'$ by holding $w$ fixed and minimizing the related objective.

$$\arg\min_{x'} \max_\lambda \lambda(f_w(x') - y')^2 + d(x_i, x')$$

Equation 2

Where $d(\cdot, \cdot)$ is a distance function that measures how far the counterfactual $x'$ and the original data point $x_i$ are from one another. In practice, maximisation over $\lambda$ is done by iteratively solving for $x'$ and increasing $\lambda$ until a sufficiently close solution is found.

The choice of optimiser for these problems is relatively unimportant. In practice, any optimiser capable of training the classifier under Equation 1 seems to work equally well, and we use ADAM[62] for all experiments. As local minima are a concern, we initialise each run with different random values for $x'$ and select as our counterfactual the best minimizer of Equation 2. These different minima can be used as a diverse set of multiple counterfactuals.

---

[62] Diederik Kingma & Jimmy Ba, *ADAM: A Method for Stochastic Optimization*, ARXIV:1412.6980, at 1–4 (2014), https://arxiv.org/pdf/1412.6980.pdf [https://perma.cc/3RH4-WSXG].

Of particular importance is the choice of distance function used to decide which synthetic data point $x'$ is closest to the original data point $x_i$. As a sensible first choice, which should be refined based on subject- and task-specific requirements, we suggest use of the $L_1$ norm, or Manhattan distance, weighted by the inverse median absolute deviation. This is written as $\mathrm{MAD}_k$ for the median absolute deviation of feature $k$, over the set of points $P$:

$$\mathrm{MAD}_k = \mathrm{median}_{j \in P} \left( \, |X_{j,k} - \mathrm{median}_{l \in P}(X_{l,k})| \, \right)$$

Equation 3

We chose $d(\cdot,\cdot)$ as:

$$d(x_i, x') = \sum_{k \in F} \frac{|x_{i,k} - x'_k|}{\mathrm{MAD}_k}$$

Equation 4

This distance metric has several desirable properties. Firstly, it captures some of the intrinsic volatility of the space, which means that if a feature $k$ varies wildly across the dataset, a synthetic point $x'$ may also vary this feature while remaining close to $x_i$ under the distance metric. The use of median absolute difference rather than the more usual standard deviation also makes this metric more robust to outliers. Of equal importance are the sparsity-inducing properties of the $L_1$ norm. The $L_1$ norm is widely recognised in mathematical and machine learning circles for its tendency to induce sparse solutions in which most entries are zero when paired with an appropriate cost function.[63]

When computing human-understandable counterfactuals, this property is highly desirable as it corresponds to counterfactuals in which only a small number of variables are changed and most remain constant, making the counterfactuals much easier to communicate and comprehend. This metric works equally well on the examples we consider.

To demonstrate the importance of the choice of distance function, we illustrate below the impact of varying $d(\cdot,\cdot)$ on the LSAT dataset. A further challenge lies in ensuring that the synthetic counterfactual $x'$ corresponds to a valid data point. We illustrate some of the pitfalls and remedies for dealing with discrete features when computing counterfactuals.

---

[63] *See, e.g.*, Emmanuel J. Candes, Justin K. Romberg & Terence Tao, *Stable Signal Recovery from Incomplete and Inaccurate Measurements*, 59 COMM. PURE & APPLIED MATHEMATICS 1207, 1212 (2006).

### A. LSAT DATASET

We first consider the generation of counterfactuals on the LSAT[64] dataset. In particular, we consider a stripped-down version used in the fairness literature[65] that attempts to predict students' first-year average grade on the basis of their race, grade-point average prior to law school, and law school entrance exam scores. This stripped-down version of the LSAT dataset is used in the fairness literature, as classifiers trained on this data naturally exhibit bias against 'black' people.[66] As a result, we will find evidence of this bias in our neural network in some of the counterfactuals we generate.

We generate a three-layer fully-connected neural-network, with two hidden layers of 20 neurons each feeding into a final classifier. Even a small model like this has 941 different weights controlling its behaviour and 40 neurons that exhibit complex interdependencies, which makes conveying its internal state challenging.

Choosing $d$ as the unweighted squared Euclidean distance

$$d(x_i, x') = \sum_{k \in F} (x_{i,k} - x'_k)^2$$

Equation 5

we consider the Counterfactual, "What would have to be changed to give a predicted score of 0?"[67] Directly solving for Eq. 2 gives the results in the central block labelled "Counterfactuals" in Table 1.

| | | | | Unnormalised L2 | | | | | |
|---|---|---|---|---|---|---|---|---|---|
| | Original Data | | | | Counterfactuals | | | Counterfactual Hybrid | |
| score | GPA | LSAT | Race | GPA | LSAT | Race | GPA | LSAT | Race |
| 0.17 | 3.1 | 39.0 | 0 | 3.0 | 39.0 | 0.3 | 1.5 | 38.4 | 0 |
| 0.54 | 3.7 | 48.0 | 0 | 3.5 | 47.9 | 0.9 | -1.6 | 45.9 | 0 |
| -0.77 | 3.3 | 28.0 | 1 | 3.5 | 28.1 | -0.3 | 5.3 | 28.9 | 0 |
| -0.83 | 2.4 | 28.5 | 1 | 2.6 | 28.6 | -0.4 | 4.8 | 29.4 | 0 |
| -0.57 | 2.7 | 18.3 | 0 | 2.9 | 18.4 | -1.0 | 8.4 | 20.6 | 0 |

*Table 1 - Unnormalized L2*

---

Two artefacts are immediately apparent. The first is that although in this dataset, race is modelled using a discrete variable that can only take the labels 0 or 1, corresponding to 'white' or 'black' respectively, a variety of meaningless values, either fractional or negative, have been assigned to it. In the literature on adversarial perturbation, generally values are capped to lie within a sensible range such as [0,1] to stop some of these artefacts from occurring. However, this would still allow the fractional solutions shown in the bottom two examples. Instead, we clamp the race variable forcing it to take either value 0 or 1 in two separate run-throughs, and then take as a solution the closest counterfactual found in either of the runs. These results can be seen in the rightmost column "Counterfactual Hybrid." The algorithm now suggests always changing the race to 'white' as part of the counterfactual. Of particular note is that the counterfactuals show that 'black' students would get better scores.

The second artefact is that the algorithm much prefers significantly varying the GPA than the exam results, and this is down to our choice of distance function. We took as $d(\cdot,\cdot)$, the squared Euclidean distance, and this generally prefers changes that are as small as possible and spread uniformly across all variables. However, the range of the GPA is much smaller than that of the exam scores. Adjusting for this by normalising each component by its standard deviation, i.e.

$$d(x_i, x') = \sum_{k \in F} \frac{(x_{i,k} - x'_k)^2}{\mathrm{std}_{j \in P}(x_{j,k})}$$

Equation 6

gives the set of counterfactuals shown in Table 2.

| | | | | Normalised L2 | | | | | |
| score | GPA | LSAT | Race | GPA | LSAT | Race | GPA | LSAT | Race |
|---|---|---|---|---|---|---|---|---|---|
| 0.17 | 3.1 | 39.0 | 0 | 3.0 | 37.0 | 0.2 | 3.0 | 34.0 | 0 |
| 0.54 | 3.7 | 48.0 | 0 | 3.5 | 39.5 | 0.4 | 3.5 | 33.1 | 0 |
| -0.77 | 3.3 | 28.0 | 1 | 3.5 | 39.8 | 0.4 | 3.4 | 33.4 | 0 |
| -0.83 | 2.4 | 28.5 | 1 | 2.7 | 37.4 | 0.2 | 2.6 | 35.7 | 0 |
| -0.57 | 2.7 | 18.3 | 0 | 2.8 | 28.1 | -0.4 | 2.9 | 34.1 | 0 |

*Table 2 - Normalised L2*

After normalisation, the GPA remains much more consistent, and naturally remains within an expected range of values. Note that for 'black' students, race does vary under the computed counterfactual, revealing a dependence between the decision and race (which is often a legally protected attribute).

Finally, we show the use of the $L_1$ norm weighted by the inverse median absolute deviation (Table 3). This returns similar but sparser results to the weighted squared Euclidean distance, with the GPA not being changed under the counterfactuals.

| | | | | Normalised L1 | | | | | |
|---|---|---|---|---|---|---|---|---|---|
| | Original Data | | | Counterfactuals Continuous | | | Counterfactual Hybrid | | |
| score | GPA | LSAT | Race | GPA | LSAT | Race | GPA | LSAT | Race |
| 0.17 | 3.1 | 39.0 | 0 | 3.1 | 35.0 | 0.1 | 3.1 | 34.0 | 0 |
| 0.54 | 3.7 | 48.0 | 0 | 3.7 | 33.5 | 0.0 | 3.7 | 32.4 | 0 |
| -0.77 | 3.3 | 28.0 | 1 | 3.3 | 34.4 | 0.1 | 3.3 | 33.5 | 0 |
| -0.83 | 2.4 | 28.5 | 1 | 2.4 | 39.3 | 0.2 | 2.4 | 35.8 | 0 |
| -0.57 | 2.7 | 18.3 | 0 | 2.7 | 35.8 | 0.1 | 2.7 | 34.9 | 0 |

Table 3 - Normalised L1

These final Normalised $L_1$ Hybrid Counterfactuals can be expressed in a more accessible text form that only describes the alterations to the original data:

> **Person 1:** If your LSAT was 34.0, you would have an average predicted score (0).
>
> **Person 2:** If your LSAT was 32.4, you would have an average predicted score (0).
>
> **Person 3:** If your LSAT was 33.5, and you were 'white', you would have an average predicted score (0).
>
> **Person 4:** If your LSAT was 35.8, and you were 'white', you would have an average predicted score (0).
>
> **Person 5:** If your LSAT was 34.9, you would have an average predicted score (0).

### B. PIMA DIABETES DATABASE

To demonstrate Counterfactuals on a more complex problem, we consider a database used to predict whether women of Pima heritage are at risk of diabetes.[68] We generate a classifier that returns a risk score between [0, 1] by training a similar three-layer fully-connected neural-network with two hidden layers of 20 neurons to perform logistic regression. This classifier takes as input 8 different variables of varying predictive power, including number of pregnancies, age and BMI. Counterfactuals are generated to answer the question "What would have to be different for this individual to have a risk score of 0.5?" To induce sparsity in the answer and generate counterfactuals that are easy for a human to evaluate, with only a small number of changed variables, we make use of the $L_1$ norm, or Manhattan distance, weighted by the inverse median absolute deviation, instead of the Euclidean distance. We also cap variables to prevent them from going outside the range seen in the training data.

---

[68] Jack W. Smith et al., *Using the ADAP Learning Algorithm to Forecast the Onset of Diabetes Mellitus*, PROC. ANN. SYMP. ON COMPUT. APPLICATION MED. CARE 261, 261–62 (1988).

With this done, the counterfactuals typically vary from the original data only in a small number of variables, and these differences are automatically rendered in human readable text form.

> **Person 1:** If your 2-Hour serum insulin level was 154.3, you would have a score of 0.51.
> **Person 2:** If your 2-Hour serum insulin level was 169.5, you would have a score of 0.51.
> **Person 3:** If your Plasma glucose concentration was 158.3 and your 2-Hour serum insulin level was 160.5, you would have a score of 0.51.

These counterfactuals are similar to the risk factors already used by doctors to communicate, e.g. "If your body mass index is greater than 40 you are morbidly obese, and at greater risk of ill-health." However, counterfactuals may make use of multiple factors and convey a personalised risk model that takes into account other attributes that may mitigate or increase risk.

## C. CAUSAL ASSUMPTIONS AND COUNTERFACTUAL EXPLANATIONS

The reader familiar with causal modelling may have noticed that our counterfactual explanations are not making use of causal models or equivalently, that they make naive assumptions that variables are independent of one another. There are several reasons for this. One important use of counterfactual explanations is to provide the data subject with information to make a guided audit of the data and check for relevant inaccuracies in the data. Treating such errors as independent and drawn from a robust distribution such as the Laplacian (corresponding to use of the $L_1$ norm in our objective) is a sensible model for these errors. More importantly, creating and interpreting accurate causal models is difficult. Requiring data controllers to build and convey to a lay audience a causal model that accurately captures the interdependencies between measurements such as the number of pregnancies, age, and BMI is extremely challenging and may be irrelevant.

Counterfactuals generated from an accurate causal model may ultimately be of use to experts (e.g., to medical professionals trying to decide which intervention will move a patient out of an at-risk group). However, the purpose of our paper is to illustrate how far you can go with minimal assumptions and that such detailed causal models are unnecessary for counterfactual explanations to be of use.

## IV. ADVANTAGES OF COUNTERFACTUAL EXPLANATIONS

Counterfactual explanations differ markedly from existing proposals in the machine learning and legal communities (particularly regarding the GDPR's "right to explanation"),[69] while offering several advantages. Principally, counterfactuals bypass the substantial challenge of explaining the internal workings of complex machine learning systems.[70] Even if technically feasible, such explanations may be of little practical value to data subjects. In contrast, counterfactuals provide information to the data subject that is both easily digestible and practically useful for understanding the reasons for a decision, challenging them, and altering future behaviour for a better result.

The reduced regulatory burden of counterfactual explanations is also significant. Current state-of-the-art machine learning methods make decisions based upon deep networks that compose together functions more than a thousand times and with more than ten million parameters controlling their behaviour.[71] As the working memory of humans can contain around seven distinct items,[72] it remains unclear whether "human-comprehensible meaningful information" about the logic involved in a particular decision can ever exist, disregarding whether such information could be meaningfully conveyed to non-experts.[73] As such, regulations that require meaningful information regarding the internal logic to be conveyable to a lay audience could prohibit the use of many standard approaches. In contrast, counterfactual explanations do not attempt to convey the logic involved and, as shown in the previous section, are simple to compute and convey.

---

[69] Although a right to explanation is not itself legally binding, data subjects are entitled to receive "meaningful information about the logic involved, as well as the significance and the envisaged consequences" of automated decision-making under the GDPR's Art. 13–15. Wachter, Mittelstadt & Floridi, *supra* note 1, at 16. Others have proposed that these provisions require the data subject to be given information about the internal logic and the rationale of specific decisions. The information sought aligns with the type of explanation pursued in the machine learning community. For an explanation of why such information is not legally required, and why Art. 13–15 do not constitute a de facto right to explanation, see Wachter, Mittelstadt & Floridi, *supra* note 1, at 10–11, 14–19.

[70] Burrell, *supra* note 5, at 9.

[71] *See generally* Kaiming He et al., *Deep Residual Learning for Image Recognition*, PROC. IEEE CONF. ON COMPUT. VISION & PATTERN RECOGNITION 770, 770–78 (2016); Gao Huang et al., *Deep Networks with Stochastic Depth*, EUROPEAN CONF. ON COMPUT. VISION 646, 646–61 (2016).

[72] George A. Miller, *The Magical Number Seven, Plus or Minus Two: Some Limits on Our Capacity for Processing Information*, 63 PSYCHOL. REV. 81, 91 (1956).

[73] Burrell, *supra* note 5, at 9.

Such expectations of providing information regarding the internal logic of algorithmic decision-making systems have surfaced recently in relation to the GDPR and in particular, the "right to explanation." The GDPR contains numerous provisions requiring information to be communicated to individuals about automated decision-making.[74] Significant discussion has emerged in legal and machine learning communities regarding the specific requirements and limitations of the GDPR in this regard and in particular, how to provide information about decisions made by highly complex automated systems.[75] As counterfactuals provide a method to explain some of the rationale of an automated decision while avoiding the major pitfalls of interpretability or opening the "black box," they may prove a highly useful mechanism to meet the explicit requirements and background aims of the GDPR.

## V. COUNTERFACTUAL EXPLANATIONS AND THE GDPR

Although the GDPR's "right to explanation" is not legally binding, it has nonetheless connected discussion of data protection law to the longstanding question of how algorithmic decisions can be explained to experts as well as non-expert parties affected by the decision.[76] Answering this question largely depends upon the intended purpose of the explanation; the information to be provided must be tailored in terms of structure, complexity, and content with a particular aim in mind. Unfortunately, the GDPR does not explicitly define requirements for explanations of automated decision-making and provides few hints as to the intended purpose of explanations of automated decision-making.[77] Recital 71 of the GDPR, a non-binding provision and the only place where the word explanation is mentioned, states that suitable safeguards against automated decision-making should be implemented and "should include specific information to the data subject and the right to obtain human intervention, to express his or her point of view, to obtain an explanation of the decision reached after such assessment and to challenge the decision."[78]

---

[74] Regulation 2016/679 of Apr. 27, 2016, on the Protection of Natural Persons with Regard to the Processing of Personal Data and on the Free Movement of Such Data, and Repealing Directive 95/46/EC (General Data Protection Regulation) [hereinafter GDPR], GDPR, recitals 63 & 71 & arts. 13(2)(f), 14(2)(g), 15(1)(h) & 22, 2016 O.J. (L 119) 12, 14, 41, 42, 43, 46 (EU).

[75] *See generally* Wachter, Mittelstadt & Floridi, *supra* note 1; Mendoza & Bygrave, *supra* note 1; Edwards & Veale, *supra* note 1; Malgieri & Comandé, *supra* note 1; Selbst & Powles, *supra* note 1; Doshi-Velez et al., *supra* note 15; Christopher Kuner et al., *Machine Learning with Personal Data: Is Data Protection Law Smart Enough to Meet the Challenge?*, 7 INT'L DATA PRIVACY L. 1 (2017).

[76] *See* Wachter, Mittelstadt & Floridi, *supra* note 1, at 3–4.

[77] *See id.* at 42.

This is the only time where an explanation is mentioned in the GDPR, leaving the reader with little insight into what type of explanation is intended or what purpose it should serve. Based on the text, the only clear indication is that legislators wanted to clarify that some type of explanation can voluntarily be offered *after* a decision has been made. This can be seen as Recital 71 separates "specific information" which should be given before a decision is made,[79] from safeguards that apply *after* a decision has been made[80] ("an explanation of the decision *reached after such assessment*" (emphasis added)).[81] Further indications are not provided of the intended content of such ex post explanations.[82]

The content of an explanation must reflect its intended purpose. Given the lack of guidance in the GDPR, many aims for explanations are feasible. Reflecting the GDPR's emphasis on protections and rights for individuals,[83] here we examine potential purposes for explanations from the perspective of the data subject. We propose three possible aims of explanations of automated decisions: to enhance understanding of the scope of automated decision-making and the reasons for a particular decision, to help contest a decision, and to alter future behaviour to potentially receive a preferred outcome. This is not an exhaustive list of potential aims of explanations, but rather reflects how the recipient of an automated decision, as with any type of decision, may wish to understand its scope, effects, and rationale and take actions in response. In the following sections, we assess how these three purposes are reflected in the GDPR and the extent to which counterfactual explanations meet and exceed the GDPR's requirements.

---

[78] Regulation 2016/679, GDPR, recital 71, 2016 O.J. (L 119) 14 (EU).

[79] Jörg Hladjk, *DS-GVO Art. 22 Automatisierte Entscheidungen im Einzelfall einschließlich Profiling*, *in* DATENSCHUTZ-GRUNDVERORDNUNG 529, 535 (Eugen Ehmann & Martin Selmayr eds., 1st ed. 2017).

[80] The European Parliament makes the same distinction (information obligations vs. explanations of automated decisions) in their draft report on civil law rules on robotics when referring to the GDPR. *See* European Parliament Committee on Legal Affairs, *Draft Report with Recommendations to the Commission on Civil Law Rules on Robotics* (Mar. 31, 2016), http://www.europarl.europa.eu/sides/getDoc.do?pubRef=-//EP//NONSGML%2BCOMPARL%2BPE-582.443%2B01%2BDOC%2BPDF%2BV0//EN [https://perma.cc/A2L8-FKMP]; Sandra Wachter, Brent Mittelstadt & Luciano Floridi, *Transparent, Explainable, and Accountable AI for Robotics*, 2 SCI. ROBOTICS 1, 1 (2017); Hladjk, *supra* note 79 at 535–36 (supporting this view that an explanation should be given after a decision has been taken, while recognising that this is not legally binding).

[81] Regulation 2016/679, GDPR, recital 71, 2016 O.J. (L 119) 14 (EU).

[82] *See generally* European Commission, *Proposal for a Regulation of the European Parliament and the Council on the Protection of Individuals with Regard to the Processing of Personal Data and On the Free Movement of Such Data (General Data Protection Regulation)*, COM (2012) 11 final (Jan. 25, 2012), http://ec.europa.eu/justice/data-protection/document/review2012/com_2012_11_en.pdf [https://perma.cc/6QSX-F4JX].

[83] Christopher Kuner, *The European Commission's Proposed Data Protection Regulation: A Copernican Revolution in European Data Protection Law*, PRIVACY & SEC. L. REP. 1, 6–7, 8 (2012).

## A. EXPLANATIONS TO UNDERSTAND DECISIONS

One potential purpose of explanations is to provide the data subject with understanding of the scope of automated decision-making, and the reasons that led to a particular decision. Several provisions in the GDPR can support a data subject's understanding of automated decision-making, although the types of information that must be shared tend to enhance a broad understanding of automated decision-making systems, as opposed to the rationale of specific decisions.[84] As a result, the GDPR does not appear to require opening the "black box" to explain the internal logic of the decision-making system to data subjects. With this in mind, counterfactuals can provide information aligned with the GDPR's various informational requirements, while also providing some insight into the reasons that led to a particular decision. Counterfactuals, thus, could meet and exceed the requirements of the GDPR.

The description of explanations in Recital 71 does not include a requirement to open the "black box."[85] Understanding the internal logic

---

[84] *See* Wachter, Mittelstadt & Floridi, *supra* note 1, at 5.

[85] *Id.; see also* ARTICLE 29 DATA PROTECTION WORKING PARTY, GUIDELINES ON AUTOMATED INDIVIDUAL DECISION-MAKING AND PROFILING FOR THE PURPOSES OF REGULATION 2016/679 29 (2018), http://ec.europa.eu/newsroom/article29/document.cfm?doc_id=49826. The Guidelines, which are very ambiguous, seem to support the claim that such a requirement is not only absent, but also might not have been intended. On the one hand transparency in how decisions are made (Recital 71) appears to be very important. *See* Article 29 Data Protection Working Party, *Guidelines* at 27. However, at the same time, the guidelines state that the aim of Art. 15(1)(h) is not to create individual explanations that require understanding the internal logic of the algorithm. *Id.* at 27. Hence, the guidelines suggest that Art. 15(1)(h) calls for information about general system functionality, as is the case with its counterparts in Art. 13(2)(f) and Art. 14(2)(g). This reading of Articles 13–15 would suggest that the Article 29 Working Party does not view non-binding Recital 71 as a requirement to explain the internal logic of individual decisions, as even the legally

of the algorithmic decision-making system is not explicitly required. Elsewhere, the GDPR contains transparency mechanisms,[86] notification duties,[87] and the right of access,[88] all of which create informational requirements concerning automated decision-making. Art. 13–15 describe what kind of information needs to be provided if data are collected, either immediately when collected from the data subject,[89] the latest after a month when collected from a third party,[90] or at any time if requested from the data subject.[91] Among other things, Art. 12 explains how this information (as defined in Art. 13–14) should be conveyed.[92] Art. 12–14 suggest that data subjects must be provided with "a meaningful overview of the intended processing,"[93] including "the existence of automated decision-making, including profiling, referred to in Art. 22(1) and (4) and, at least in those cases, meaningful information about the logic involved, as well as the significance and the envisaged consequences of such processing for the data subject,"[94] as opposed to a detailed explanation of the internal logic of a system after a decision has been made.[95] Rather they aim to offer a generic overview of intended processing activities, which enhances the data subject's understanding of the scope and purpose of automated decision-making.[96]

---

binding text in Article 15(1)(h), which is sufficiently vague to allow such an interpretation, *see* Wachter, Mittelstadt & Floridi, *supra* note 1, is not thought to create such a requirement. For further support that Recital 71 does not hinge on opening the black box, see Martini, *DS-GVO Art. 22 Automatisierte Entscheidungen im Einzelfall einschließlich Profiling, in* DATENSCHUTZ-GRUNDVERORDNUNG Rn 35-37 (Paal & Pauly eds., 1st ed. 2017).

[86] Regulation 2016/679, GDPR, art. 12, 2016 O.J. (L 119) 39–40 (EU).

[87] Regulation 2016/679, GDPR, arts. 13 & 14, 2016 O.J. (L 119) 40–42 (EU).

[88] Regulation 2016/679, GDPR, art. 15, 2016 O.J. (L 119) 43 (EU).

[89] Regulation 2016/679, GDPR, art. 13–15, 2016 O.J. (L 119) 40–43 (EU).

[90] Regulation 2016/679, GDPR, art. 14, 2016 O.J. (L 119) 41–42 (EU).

[91] Regulation 2016/679, GDPR, art. 15, 2016 O.J. (L 119) 43 (EU).

[92] Dirk Heckmann & Anne Paschke, *DS-GVO Art. 12 Transparente Information, Kommunikation, in* DATENSCHUTZ-GRUNDVERORDNUNG 367, 370 (Eugen Ehmann & Martin Selmayr eds., 1st ed. 2017).

[93] Regulation 2016/679, GDPR, art. 12(7), 2016 O.J. (L 119) 40 (EU).

[94] Regulation 2016/679, GDPR, arts. 13(2)(f) & 14(2)(g), 2016 O.J. (L 119) 41, 42 (EU).

[95] Lorenz Franck, *DS-GVO Art. 12 Transparente Information, Kommunikation*, in DATENSCHUTZ-GRUNDVERORDNUNG VO (EU) 2016/679 316, 320 (Peter Gola ed., 1st ed. 2017); Sebastian Schulz, *DS-GVO Art. 22 Automatisierte Entscheidungen im Einzelfall, in* DATENSCHUTZ-GRUNDVERORDNUNG VO (EU) 2016/679 410, 418–19 (Peter Gola ed., 1st ed. 2017); Suzanne Rodway, *Just How Fair Will Processing Notices Need to Be Under the GDPR*, 16 PRIV. & DATA PROT. 16, 16–17 (2016).

[96] *See* Kuner, *supra* note 75, at 2; ROSEMARY JAY, GUIDE TO THE GENERAL DATA PROTECTION REGULATION: A COMPANION TO DATA PROTECTION LAW AND PRACTICE 226 (4th Revised ed. 2017).

Art. 12(7) clarifies that the aim of Art. 13-14 is to provide "in an easily visible, intelligible and clearly legible manner, a *meaningful overview of the intended processing*."[97] Two requirements are notable: (1) that the information provided must be meaningful to its recipient and broad in scope (a "*meaningful overview*"), and (2) that the notification occurs prior to processing ("*intended* processing").

To understand what would constitute a meaningful overview, the envisioned medium of disclosure is instructive. Broadly applicable information appears to be required, rather than personalised disclosures. Legal scholars have suggested that notification duties can be satisfied via updates to existing privacy statements or notices[98] (e.g. those displayed on websites or using QR codes).[99] This requirement does not change based on the form of data collection.[100] When data are collected from a third party,[101] an email sent to the data subject linking to the data controller's privacy statement(s) could suffice.[102] The same holds true for personalised links[103] referring to the privacy notice. Tools similar to those currently used to make users aware of the usage of cookies or monitoring shopping behaviour can be envisioned to satisfy the requirements in Art. 14, thus making data subjects immediately aware of data collection.[104] Detailed information appears to not be necessary as Art. 12(7) states that the required information can be provided along with standardised icons.[105] In trilogue, the European Parliament proposed several standardised icons that were ultimately not adopted (see Appendix 2). Despite this, the proposed icons reveal the initial expectations of regulators for simple, easily understood information.[106]

---

[97] Regulation 2016/679, GDPR, art. 12(7), 2016 O.J. (L 119) 40 (EU) (emphasis added).

[98] *See, e.g.,* ALAIN BENSOUSSAN, GENERAL DATA PROTECTION REGULATION: TEXTS, COMMENTARIES AND PRACTICAL GUIDELINES 113 (1st ed. 2017); Franck, *supra* note 95, at 320; JAY, *supra* note 96 at 223; Heckmann & Paschke, *supra* note 92, at 375–76; Rainer Knyrim, *DS-GVO Art. 14 Informationspflicht bei Erhebung von Daten*, *in* DATENSCHUTZ-GRUNDVERORDNUNG 412, 417–18 (Eugen Ehmann & Martin Selmayr eds., 1st ed. 2017).

[99] Lorenz Franck, *DS-GVO Art. 13 Informationspflicht bei Erhebung von Daten*, *in* DATENSCHUTZ-GRUNDVERORDNUNG VO (EU) 2016/679 331, 338–39 (Peter Gola ed., 1st ed. 2017).

[100] Article 29 Data Protection Working Party, *supra* note 85, at 25-6.

[101] Regulation 2016/679, GDPR, art. 14, 2016 O.J. (L 119) 41 (EU).

[102] Knyrim, *supra* note 98, at 415–19.

[103] Franck, *supra* note 95, at 322–23.

[104] Knyrim, *supra* note 98, at 420.

[105] *Id.* at 417; ARTICLE 29 DATA PROTECTION WORKING PARTY, GUIDELINES ON TRANSPARENCY UNDER REGULATION 2016/679 (2017).

[106] The European Commission is tasked in Art. 12(8) to develop such icons.

These examples suggest Art. 13–14 aim to provide a general overview of data processing that will be meaningful to all data subjects involved (e.g., all users of Twitter). The captive audience is more likely to be the general public or user base, not individual users, and their unique circumstances.[107] This format of disclosure suggests notifications should be comprehensible to a general audience with mixed expertise and background knowledge. An "uneducated layperson" may be the envisioned audience for disclosures.[108] This coincides with the general notion of Art. 12(1) that all information and communication with the data subject has to be in a "concise, transparent, intelligible and easily accessible form," suggesting in-depth technical information and 'legalese' would be inappropriate.[109] At a minimum, each provision suggests that information disclosures need to be tailored to their audience, with envisioned audiences including children and uneducated laypeople.

Notifications regarding automated decision-making[110] face particular constraints within an overall "meaningful overview." According to the Article 29 Working Party,[111] the UK Information Commissioner's Office,[112] and other commentators,[113] informing the data subject about the "significance and envisaged consequences of automated decision-making" in a very simple manner, including "how profiling might affect the data subject generally, rather than information about a specific decision" will be sufficient.[114] For instance, an explanation of how a low rating of creditworthiness can affect payment options,[115] how intended data processing may result in a credit or job application being declined,[116] or how driving behaviour might impact insurance premiums would be sufficient.[117] Similarly, "meaningful information about the logic involved" is said to require only "clarifying: of the categories of data used to create a profile; the source of the data; and why this data is considered relevant"[118] as opposed to a "detailed technical description about how an algorithm or machine learning works."[119]

---

[107] *See* Regulation 2016/679, GDPR, recital 58, 2016 O.J. (L 119) 11 (EU); Heckmann & Paschke, *supra* note 92, at 378. Note that this information can also be provided orally. *See* JAY, *supra* note 96, at 216–17 (noting this also but warning that data controllers carry the burden to prove that the information was communicated).

[108] Franck, *supra* note 95, at 322 (noting this for the elderly, uneducated people, foreigners, or children); *see also* Heckmann & Paschke, *supra* note 92, at 376–77.

[109] JAY, *supra* note 96, at 218; Heckmann & Paschke, *supra* note 92, at 376; Article 29 Data Protection Working Party, *supra* note 85, at 25-6.

[110] Regulation 2016/679, GDPR, art. 13(2), 14(2)(g), 2016 O.J. (L 119) 41–42 (EU).

[111] Article 29 Data Protection Working Party, *supra* note 85.

[112] Info. Comm'r Office, *Feedback Request - Profiling and Automated Decision-making* 15–16 (2017), https://ico.org.uk/media/about-the-ico/consultations/2013894/ico-feedback-request-profiling-and-automated-decision-making.pdf [https://perma.cc/PC33-PUS8] Note the UK's ICO is preparing new guidelines in the form of a living document, which will be continuously updated. See: UK's Information Commissioner's Office, RIGHTS RELATED TO AUTOMATED DECISION MAKING INCLUDING PROFILING (2018), https://ico.org.uk/for-organisations/guide-to-the-general-data-protection-regulation-gdpr/individual-rights/rights-related-to-automated-decision-making-including-profiling/ (last visited Mar 18, 2018).

[113] *See, e.g.,* Rodway, *supra* note 95; Paal, *DS-GVO Art. 13 Informationspflicht bei Erhebung von personenbezogenen Daten bei der betroffenen Person*, *in* DATENSCHUTZ-GRUNDVERORDNUNG (Paal & Pauly eds., 1st ed. 2017).

[114] Info. Comm'r Office, *supra* note 112, at 16.

      This view is echoed in the Article 29 Working Party's guidelines on automated individual decision-making. First the "right to explanation" is only mentioned once in the guidelines without any further details on scope or purpose. This "right" is clearly separated from the legally binding safeguards in Art. 22(3), implying that the Article 29 Working Party sees a difference in the legal standing of Recitals and legally binding provisions.[120] In fact, the guidelines does not even list the right to explanation in their "good practice suggestions" section.[121] Transparency about the fact that data controllers "are engaging in this type of activity," referring to automated decision-making, is essential and the main goal of Art. 13 and 14. The aim of these articles is thus to provide ex ante information.[122] This is also evident in the fact that the guidelines states that the phrase 'significance' and 'envisaged consequences' means "that information must be provided about intended or future processing, and how the automated decision-making might affect the data subject."[123] Elsewhere, the guidelines state that "details of the main characteristics considered in reaching the decision, the source of this information and the

---

relevance" should be provided under Art 13–14.[124] Further, the "controller should find simple ways to tell the data subject about the rationale behind, or the criteria relied on in reaching the decision. The GDPR requires the controller to provide meaningful information about the logic involved, not necessarily a complex explanation of the algorithms used or disclosure of the full algorithm."[125]

However, it must be noted that this requirement, despite referring to the decision-making rationale, seems to refer to general system functionality rather than an explanation of an individual decision.[126] The guidelines state that Art 15(1)(h), which is seen to provide identical information as Art 13(2)(f) and 14(2)(g),[127] requires the data controller to "provide the data subject with information about the envisaged consequences of the processing, rather than an explanation of a particular decision."[128] The is further supported as the guidelines state that "meaningful information about the logic involved" means that "Instead of providing a complex mathematical explanation about how algorithms or machine-learning work, the controller should consider using clear and comprehensive ways to deliver the information to the data subject, for example: the categories of data that have been or will be used in the profiling or decision-making process; why these categories are considered pertinent; how any profile used in the automated decision-making process is built, including any statistics used in the analysis; why this profile is relevant to the automated decision-making process; and how it is used for a decision concerning the data subject."[129]

Overall, according to the Article 29 Working Party, the aim of Articles 13-15 is to demonstrate how automated processes help data controllers to make more accurate, unbiased, and responsible decisions and illustrate how the data, characteristics, and method used are suitable to achieve this goal.[130] In other words, the process of decision-making and the algorithm itself do not need to be fully disclosed, but rather a description of the logic of the algorithm which may include a list of data sources or variables.[131] This position finds further support in the Working Party's guidelines on transparency,[132] which state that the notification duties in Art. 13-14 can be satisfied via standardised privacy notices, visualisation tools, and icons.

---

[124] *Id.* at 26.

[125] *Id.* at 25.

[126] For an in-depth analysis between systems functionality and rationale of a decision see Wachter, Mittelstadt & Floridi, *supra* note 1.

[127] *See* ARTICLE 29 DATA PROTECTION WORKING PARTY, *supra* note 88 at 26.

[128] *Id.* at 27. "The controller should provide the data subject with general information (notably, on factors taken into account for the decision-making process, and on their respective 'weight' on an aggregate level) which is also useful for him or her to challenge the decision" is given as an example showing that only information about system functionality will be required.

[129] ARTICLE 29 DATA PROTECTION WORKING PARTY, *supra* note 88 at 31.

[130] *See id.* at 26.

Each disclosure under Art. 13–14 must occur prior to data processing[133] or at the time of data collection, but before automated decision-making starts.[134] Evidence of this is seen in the future-oriented language used in Art. 13(2)(f) and Art. 14(2)(g),[135] the obligation for information about the necessity of providing data for processing,[136] the clarification in Art. 12(7) that information must be provided about "intended processing," the Article 29 Working Party's guidelines on transparency,[137] and other provisions and jurisprudence.[138] For automated

---

[131] Paal, *supra* note 113, at Rn. 31-32 (seeing no difference between Art 13-15 in terms what kind of information needs to be provided). *See also* Paal, *DS-GVO Art. 15 Auskunftsrecht der betroffenen Person*, *in* DATENSCHUTZ-GRUNDVERORDNUNG, Rn. 31 (Paal & Pauly eds., 1st ed. 2017). Further support is offered by the text of Recital 51 proposed by the European Parliament during Trilogue, which referred to "the general logic of the data that are undergoing the processing and what might be the consequences of such processing." European Parliament Committee on Civil Liberties, Justice and Home Affairs, *Report on the Proposal for a Regulation of the European Parliament and of the Council on the Protection of Individuals with Regard to the Processing of Personal Data and on the Free Movement of Such Data (General Data Protection Regulation)* A7-0402/2013, 21 (Nov. 21, 2013), http://www.europarl.europa.eu/sides/getDoc.do?type=REPORT&reference=A7-2013-0402&language=EN [https://perma.cc/27B4-5PWC].

[132] ARTICLE 29 DATA PROTECTION WORKING PARTY, *supra* note 108.

[133] Regulation 2016/679, GDPR, art. 12(7), 2016 O.J. (L 119) 40 (EU).

[134] *See* Wachter, Mittelstadt & Floridi, *supra* note 1, at 15; *see also* Franck, *supra* note 99; Knyrim, *supra* note 98; JAY, *supra* note 95, at 225 (arguing that the notification duties in Art. 13 need to apply *before* the data is collected); Franck, *supra* note 95 at 328 (linking this to Art. 13(2)(e) that obligates the data controllers to state "whether the provision of personal data is a statutory or contractual requirement, or a requirement necessary to enter into a contract, as well as whether the data subject is obliged to provide the personal data and of the possible consequences of failure to provide such data."). Information about the necessity to provide data must therefore be given before the data is collected. *See also* Article 29 Data Protection Working Party, *supra* note 85, at 12-3.

[135] *See* Wachter, Mittelstadt & Floridi, *supra* note 1, at 15; Frederike Kaltheuner & Elettra Bietti, *Data is power: Towards additional guidance on profiling and automated decision-making in the GDPR*, 2 J. INF. RIGHTS POLICY PRACT. (2018), https://journals.winchesteruniversitypress.org/index.php/jirpp/article/view/45 (last visited Mar 18, 2018).

[136] Regulation 2016/679, GDPR, art. 13(2)(e), 2016 O.J. (L 119) 41 (EU).

[137] ARTICLE 29 DATA PROTECTION WORKING PARTY, *supra* note 108 at 14 state that "Articles 13 and 14 set out information which must be provided to the data subject at the commencement phase of the processing cycle."

[138] *See* Rainer Knyrim, *DS-GVO Art. 13 Informationspflicht bei Erhebung von Daten*, *in* DATENSCHUTZ-GRUNDVERORDNUNG 391, 411–12 (Eugen Ehmann & Martin Selmayr eds., 1st ed. 2017); Knyrim, *supra* note 98, at 418–19 (noting that prior notification is

decision-making, it is essential that information is provided before the start, else the right not to be subject of an automated decision can never be realised. The data subject has no chance to assess the associated risks,[139] or whether one of the grounds in Art. 22(2) actually apply that allow automated decision-making. Under Art. 22(2), automated decision-making is only lawful if it "is necessary for entering into, or performance of, a contract between the data subject and a data controller,"[140] if it is authorised by Member State law,[141] or if the data subject has given explicit consent.[142] If notifications do not occur prior to processing or decision-making, data subjects would only be able to contest decisions after the fact. This can be time and cost intensive, and unable to repair financial or reputational damage. Hence, one purpose of Art. 13–14 is to make the data subject aware of future processing[143] and to allow them to decide if they want their data to be processed (e.g., consent),[144] assess the legitimacy (based on Member State law or contract), or exercise other rights in the GDPR.[145]

## 1. Broader possibilities with the right of access

The requirement for notification prior to processing applies only to the notification duties.[146] In contrast, the right of access[147] can be invoked at any time by the data subject, opening up the possibility of providing

---

also in line with the Bara and others judgment of the ECJ (C-201/14; 1.10.2015) which will have major implication for the GDPR as it shows that the court views prior notification of data transfer as essential). The ruling said that when information is gathered from a third party and transferred to another data controller for further processing (e.g. based on Member State law) prior notification of the data subject — even if no consent is required — is essential. Not least because it enables the exercise of Art. 15 (right of access) and Art. 16 (right to data rectification) as soon as data is collected.

[139] *See* Article 29 Data Protection Working Party, *supra* note 85, at 24-5.

[140] Regulation 2016/679, GDPR, art. 22(2)(a), 2016 O.J. (L 119) 46 (EU).

[141] Regulation 2016/679, GDPR, art. 22(2)(b), 2016 O.J. (L 119) 46 (EU).

[142] Regulation 2016/679, GDPR, art. 22(2)(c), 2016 O.J. (L 119) 46 (EU).

[143] Franck, *supra* note 95, at 326–28.

[144] On the importance to inform the data subject accurately (e.g., risks, safeguards, rights, consequences, etc.) to allow for informed/explicit consent, *see generally* JAY, *supra* note 96, at 218; Heckmann & Paschke, *supra* note 92, at 374–75; Schulz, *supra* note 95, at 418–19; Article 29 Data Protection Working Party, *supra* note 85, at 12-3.

[145] *See* Article 29 Data Protection Working Party, *Opinion 10/2004 on More Harmonised Information Provisions* (Nov. 25, 2004), http://www.statewatch.org/news/2004/dec/wp100.pdf [https://perma.cc/Y2DW-4F9Q] (describing the need to provide meaningful information (to raise awareness) about data collection and the need to move away from long privacy statements); *see also* Heckmann & Paschke, *supra* note 92, at 388–89.

[146] Regulation 2016/679, GDPR, arts. 13 & 14 2016 O.J. (L 119) 40–42 (EU).

[147] Regulation 2016/679, GDPR, art. 15, 2016 O.J. (L 119) 43 (EU).

information available after a decision has been made (i.e., the reasons for a specific decision). However, scholars have argued that the information supplied via notification duties and the right of access is largely identical, meaning the right of access is similarly limited in terms of the scope of "meaningful information about the logic involved as well as the significance and the envisaged consequences."[148] Information can thus largely be provided with identical tools (e.g., generic icons, privacy statements)[149] or generic templates[150] used for both notification and in response to access requests.

The narrower interpretation appears to be correct.[151] The Article 29 Working Party supports this view, explaining that the information requirements in Art. 13(2)(f), 14(2)(g), and 15(1)(h) are identical,[152] while Art. 15(1)(h) requires "that the controller should provide the data subject with information about the envisaged consequences of the processing, rather than an explanation of a particular decision."[153] A similar argument has been made by the ICO stating that Art. 13–15 aim to "provide information about how profiling might affect the data subject generally, rather than information about a specific decision."[154] Additionally, the GDPR indicates a restricted scope for the right of access when compared to Art. 13–14. Personal data of other data subjects must not be disclosed, as this could infringe their privacy. Access requests can also contravene trade secrets or intellectual property rights (Art. 15(4) and Recital 63), meaning an appropriate balance between the data subject and controller's interests must be struck.[155]

---

[148] Regulation 2016/679, GDPR, art. 15(1)(h), 2016 O.J. (L 119) 43 (EU).

[149] Paal, *supra* note 131, at Rn. 31; Franck, *supra* note 95, at 328; Lorenz Franck, *DS-GVO Art. 15 Auskunftsrecht der betroffenen Person*, *in* DATENSCHUTZ-GRUNDVERORDNUNG VO (EU) 2016/679 348, 349–50 (Peter Gola ed., 1st ed. 2017); Heckmann & Paschke, *supra* note 92, at 382–83.

[150] Franck, *supra* note 95, at 320 argues templates will be helpful, because as soon as Art. 15 is lodged data controllers have to inform about all the information in Art. 15, regardless of the actual request. *See also* Ehmann, *supra* note 119, at 431–32.

[151] *See* Wachter, Mittelstadt & Floridi, *supra* note 1; Michael Veale & Lilian Edwards, *Clarity, Surprises, and Further Questions in the Article 29 Working Party Draft Guidance on Automated Decision-Making and Profiling*, COMPUT. L. & SECURITY REV. (forthcoming 2018) (manuscript at 3), https://papers.ssrn.com/abstract=3071679 [https://perma.cc/Z37Z-TM3W].

[152] Article 29 Data Protection Working Party, *supra* note 85, at 26-7.

[153] *Id.* at 27.

[154] The intention of Art. 15 is to provide a control mechanism for data subjects to request at any time more or less the same information as Art. 13–14, without having to rely on legal compliance with the notification duties by data controllers. Ehmann, *supra* note 119, 426–427; Info. Comm'r's Office, *supra* note 112, at 16.

[155] Franck, *supra* note 149, 355; Ehmann, *supra* note 119, at 434–435.

## 2. Understanding through counterfactuals

Counterfactual explanations meet and exceed the aims and requirements of the GDPR's transparency mechanisms[156], notification duties[157], and right of access[158], which provide data subjects with information to understand the scope of automated decision-making. As argued above, Recital 71 does not give any clear indication of the intended purpose or content of explanations, including whether the internal logic of the algorithm must be explained. By providing simple "if-then" statements, counterfactuals align with the requirement to communicate information to data subjects in a "concise, transparent, intelligible and easily accessible form."[159] They simultaneously provide greater insight into the data subject's personal situation and the reasons behind relevant automated decisions than an overview tailored to a general audience. Counterfactuals are also less likely to infringe on trade secrets or the rights and freedoms of others (e.g., privacy), since no data of other data subjects or detailed information about the algorithm needs to be disclosed, in line with restrictions on the right of access.[160]

Perhaps most importantly, counterfactuals offer an explanation of some of the rationale of specific automated decisions, without needing to explain the internal logic of how a decision was reached (beyond a specific, limited set of dependencies between variables and the decision). This type of information is in line with the guidance mentioned above from the Article 29 Working Party[161] and the UK's ICO.[162] While opening the black box is not legally required, some information about the "logic involved" in automated decision-making must be provided.[163] Under the Data Protection Directive's right of access, disclosing the algorithm's source code, formula, weights, full set of variables, and information about reference groups has generally not been required.[164] The GDPR's right of access is likely to present similar requirements. Counterfactuals largely follow this precedent by disclosing only the influence of select external facts and variables on a specific decision. Although Art. 13(2)(f), Art. 14(2)(g), and Art. 15(1)(h) do not require information about specific decisions,[165] counterfactuals represent a minimal form of disclosure to inform the data subject about the "logic involved" in specific decisions. This form of disclosure regulatory burden for data controllers is minimised, as resolving the technical difficulties of interpretability or explaining the internal logic of complex systems to non-experts is not required to compute and communicate counterfactual explanations. Counterfactuals can thus be recommended as a minimally burdensome and disruptive technique to help data subjects understand the rationale of specific decisions beyond the explicit legal requirements of Art. 13(2)(f), Art. 14(2)(g), and Art. 15(1)(h).

---

[156] Regulation 2016/679, GDPR, art. 12, 2016 O.J. (L 119) 39–40 (EU).
[157] Regulation 2016/679, GDPR, arts. 13 & 14, 2016 O.J. (L 119) 40–42 (EU).
[158] Regulation 2016/679, GDPR, art. 15, 2016 O.J. (L 119) 43 (EU).
[159] Regulation 2016/679, GDPR, art. 12(1), 2016 O.J. (L 119) 39 (EU).
[160] *See* Regulation 2016/679, GDPR, recital 63 & art. 15(4), 2016 O.J. (L 119) 12, 43 (EU).
[161] *See generally* Article 29 Data Protection Working Party, *supra* note 85.
[162] *See generally* Info. Comm'r's Office, *supra* note 112.
[163] *See* Regulation 2016/679, GDPR, arts. 13(2)(f), 14(2)(g) & 15(1)(h), 2016 O.J. (L 119) 41–43 (EU).
[164] For an in-depth analysis of this jurisprudence, see generally Wachter, Mittelstadt & Floridi, *supra* note 1.

## B. EXPLANATIONS TO CONTEST DECISIONS

Another possible purpose of explanations is to provide information that helps contest automated decisions when an adverse or otherwise undesired decision is received. A right to contest decisions is provided as a safeguard against automated decision-making in Art. 22(3).

Contesting a decision can aim to reverse or nullify the decision and return to a status where no decision has been made, or to alter the result and receive an alternative decision. If the reasons that led to a decision need to be explained, the affected party can assess whether these reasons were legitimate and contest the assessment as required.

How a decision can be contested depends on whether the safeguards in Art. 22(3) (i.e., rights to obtain human intervention, express views, and contest the decision) are interpreted as a unit that must be invoked together, or as individual rights that can be invoked separately or in any possible combination.[166] To gauge the scope of explanations according to their purpose and aim, different possible models for contesting an automated decision need to be assessed.

Four models are possible. If the safeguards are a unit and must be invoked together, it is likely that some human involvement is necessary to issue a new decision. This could either be a human making the decision without any algorithmic help, hence the new result is a human decision rather than an automated decision. Alternatively, a person could be required to make a decision taking the algorithmic assessment and/or the data subject's objections into account, which would be human assessment with algorithmic elements. In both cases data subjects would lose their safeguards against the subsequent decision, as both types of decision are not based "solely on automated processing" and thus do not meet the definition of automated individual decisions in Art. 22(1).[167] Another possibility is that a person could be required to monitor the input data and processing (e.g., based on the data subject's objections), with a new decision made solely by the algorithmic system. In this case the Art. 22(3) safeguards still apply to the new decision.[168] Finally, if the safeguards can be separated, and data subjects can invoke their right to contest the decision without invoking their right to obtain human intervention or express their views, a new decision could be issued with no human involvement. This decision could be contested again under Art. 22(3). It is unclear which of these models will be preferred following implementation of the GDPR.[169]

---

[165] *See* Article 29 Data Protection Working Party, *supra* note 85, at 26-7.
[166] *See* Martini, *supra* note 85.

The question remains what explanations would be helpful to contest decisions. This will depend on the contesting model. The first model where a human makes a new decision and disregards everything the algorithm suggested, an explanation of the rationale of the original decision could be informative, but will not practically impact the new decision made entirely by a human decision-maker. For each of the other models, where algorithmic involvement is envisioned, an explanation of the rationale of the decision could be helpful to identify potential grounds for contesting, such as inaccuracies in the input data, problematic inferences, or other flaws in the algorithmic reasoning.[170]

---

[167] Regulation 2016/679, GDPR, art. 22, 2016 O.J. (L 119) 46 (EU).

[168] However, if the new automated decision is communicated to the data subject by a person, it may not be considered "solely automated" and not subject to the Art 22(3) safeguards. The precise limitations on "solely automated" in Art 22(1) remain unclear. *See also* Article 29 Data Protection Working Party, *supra* note 85, at 10 (explaining that fabricated human involvement should not be used as a loophole).

[169] Either interpretation is possible, as indicated by the European Parliament's proposal to add the following text to Article 20 in an earlier draft of the GDPR, "[t]he suitable measures to safeguard the data subject's legitimate interests referred to in paragraph 2 shall include the right to obtain human assessment and an explanation of the decision reached after such assessment." This text clarified that a human would need to assess the decision in question. However, this text was not adopted in the end, which leaves implementation of any of the four models possible. With that said, treating the safeguards as individually enforceable may be the most sensible option. Individuals can have an interest in expressing their views or obtaining human intervention when a decision is poorly understood or misunderstood. Both interests do not, however, necessarily lead to challenging the decision, particularly if challenges are costly or have a low likelihood of success.

Even though an explanation of the rationale of a decision could be helpful to contest decisions, it does not imply an explanation is required by the GDPR or is the intended aim of the non-binding right to explanation.[171] Recital 71 does not specify the aim of the right or what information should be revealed, and does not explicitly require the algorithm's internal logic to be explained. An explicit link is not established in the GDPR between the right to explanation and the right to contest, wherein the former would provide information necessary to exercise the latter.[172] Further, there is no reason to assume that the safeguards in Art. 22(3) must be exercised together, rather than independently of one another. Therefore, explanations under Recital 71 are not a necessary precondition to contest unfavourable decisions, even though this might be helpful.

Similarly, an explicit link has not been made between the right to contest and the transparency mechanisms,[173] notification duties,[174] right of access,[175] meaning the information provided through these rights and duties need not be explicitly tailored to help data subjects successfully contest decisions.[176]

Nonetheless, information provided by Art. 12–15 may be helpful for contesting. Support is evident in the fact that notification duties aim to facilitate the exercise of other rights in the GDPR to increase individual control over personal data processing.[177] To achieve this, Art. 13(2)(b), 14(2)(c), and 15(1)(e) obligate data controllers to inform data subjects about their rights in Art. 15–21[178] at the time when the data is collected,[179] within one month when obtained from a third party,[180] or at any time if requested by the data subject.[181] However, Art. 22 appears not to be covered by these provisions due to the odd phrasing of the obligation to inform of "the existence of automated decision-making, including profiling, referred to in Art. 22(1) and (4) and, at least in those cases, meaningful information about the logic involved, as well as the significance and the envisaged consequences of such processing for the data subject."[182]

---

[170]  Brent Mittelstadt et al., *The Ethics of Algorithms: Mapping the Debate*, Big Data Soc. (2016), http://bds.sagepub.com/lookup/doi/10.1177/2053951716679679 [https://perma.cc/YB4Y-9MXD].

[171]  *See* Regulation 2016/679, GDPR, recital 71, 2016 O.J. (L 119) 14 (EU).

[172]  *See* Hladjk, *supra* note 79, at 535–536; Schulz, *supra* note 95, 419–420 (arguing that "contesting" and "explaining" the decision are separate and independent safeguards).

[173]  Regulation 2016/679, GDPR, art. 12, 2016 O.J. (L 119) 39–40 (EU).

[174]  Regulation 2016/679, GDPR, arts. 13 & 14, 2016 O.J. (L 119) 40–42 (EU).

[175]  Regulation 2016/679, GDPR, art. 15, 2016 O.J. (L 119) 43 (EU).

[176]  At the same time, the Article 29 Data Protection Working Party argues that transparency in processing is essential to contesting, and that the reasons for the decisions and the legitimate basis should be known. *See* Article 29 Data Protection Working Party, *supra* note 85, at 27. However, the guidelines leave open whether this requires opening the black box and disclosing the algorithm. *See id.* This seems unlikely as the guidelines state that not even Art. 15(1)(h) aims to offer an explanation about an individual decision. *See id.* at 27. Hence it can be assumed that the information provided does not need to include an explanation of the internal logic of a specific decision.

[177]  *See* Regulation 2016/679, GDPR, art. 12(2), 2016 O.J. (L 119) 40 (EU).

        As argued above, Art. 13–15 will provide a meaningful overview of automated decision-making tailored to a general audience. On the surface, such an overview is not immediately useful for contesting decisions, as information about the rationale of individual decisions is not provided. In describing information to be provided about automated decision-making, these Articles explicitly refer only to Art. 22(1) and (4). It follows that data subjects do not need to be informed about the safeguards against automated decision-making such as the right to contest.[183] This limitation is telling. If the aim of Art. 13–15 were to facilitate contesting decisions by providing useful, individual-level information, one would expect the right to contest or Art. 22(3) to be explicitly discussed. Similarly, Art. 13–15 seem not require to inform the data subject about their right not to be subject to an automated individual decision,[184] from which a right to contest decisions could be inferred.[185]

---

[178] *See* Regulation 2016/679, GDPR, art. 15, 2016 O.J. (L 119) 43 (EU). If invoked via Art. 15(1)(e), data controllers only have to inform about the rights enshrined in Art. 16, 17, 18, 19, and 21.

[179] Regulation 2016/679, GDPR, art. 13, 2016 O.J. (L 119) 40–41 (EU).

[180] Regulation 2016/679, GDPR, art. 14(3)(a), 2016 O.J. (L 119) 42 (EU).

[181] Regulation 2016/679, GDPR, art. 15, 2016 O.J. (L 119) 43 (EU).

[182] Regulation 2016/679, GDPR, art. 15(1)(h), 2016 O.J. (L 119) 43 (EU).

[183] *See* Regulation 2016/679, GDPR, art. 22(3), 2016 O.J. (L 119) 46 (EU); Schulz, *supra* note 95, 419–420 (arguing that data controllers only have to inform about the safeguards after an adverse decision has been issued). In fact, Art. 12(3) introduces a very complicated model where the data controller has to inform upon request what kind of measures have been taken to satisfy a request under Art. 15–22, without being informed about the safeguards beforehand. *See also*, Martini, *supra* note 85, at Rn. 39–40 (also acknowledging this loophole).

[184] *See* Regulation 2016/679, GDPR, art. 22(1), 2016 O.J. (L 119) 46 (EU).

[185] It is important to note that Art. 13–15 obligate to inform about the general right to object to processing (which forces data controllers to stop processing) in Art. 21. Together with the information about the legitimate basis for processing provided in Art. 13(1)(c) and Art. 14(1)(c), this information could be used to contest decisions. However, this arrangement may place an unreasonable burden on the data subject. Contesting should be made as easy as possible, not least because the chances of successfully forcing the data controller to stop processing under Art. 21 are different from under Art. 22. For

In fact, in an earlier draft of the GDPR it was suggested that the information rights should refer to Art. 20 as a whole.[186] Ultimately, this approach was not adopted, suggesting that the lack of useful information for contesting decisions was intentional.

This lack of an explicit link to the safeguards against automated decision-making is in many ways unsurprising. Art. 12–15 aim to inform data subjects about the existence of their rights in the GDPR,[187] and to facilitate their exercise.[188] This does not, however, mean that the controller is required to provide other information to help the data subject to exercise her rights.[189] Rather, the data subject only needs to be informed about the existence of her rights, and provided with the necessary infrastructure for their exercise[190] (e.g., web portals for complaints), including the elimination of unnecessary bureaucratic hurdles,[191] a guarantee of reasonable response time to queries lodged[192] as stated in Art. 12(3), and the opportunity to interact with someone who has the power to change the decision.[193] However, the data subject remains responsible to exercise her rights independently.[194] As one commentator notes, Art. 15 does not create a duty to legal consultancy;[195] rather, it is sufficient that the data controllers inform about the existing rights in the GDPR. Unfortunately, Recital 60, which vaguely states that "any further information necessary to ensure fair and transparent processing taking into account the specific circumstances and context in which the personal data are processed" should be provided to the data subject, does not offer

---

example, legitimate interest of data controller can trump a data subject right to object to data processing under Art. 21. However, Art. 22 does not allow automated decision-making on the basis of legitimate interest of the data controller (only explicit consent, law, or contract). This information will be useful for a data subject if they want to prevent data controllers from making decisions or to contest decisions. *See* Regulation 2016/679, GDPR, arts. 13–15 & 21–22, 2016 O.J. (L 119) 40–43, 45–46 (EU).

[186] *See* EUROPEAN DIGITAL RIGHTS, COMPARISON OF THE PARLIAMENT AND COUNCIL TEXT ON THE GENERAL DATA PROTECTION REGULATION 131 (2016), https://edri.org/files/EP_Council_Comparison.pdf [https://perma.cc/CY6H-P97Z].

[187] *See* Knyrim, *supra* note 98, 415–416.

[188] *See* Heckmann & Paschke, *supra* note 92, at 371–372; Ehmann, *supra* note 119, at 425.

[189] *See* Heckmann & Paschke, *supra* note 92, at 378–379 (articulating that easily understood information provided about data subject rights in Art. 15–22 is sufficient to facilitate their exercise).

[190] *See* Bensoussan, *supra* note 98, at 114.

[191] *See* Heckmann & Paschke, *supra* note 92, at 379–380.

[192] *See* Franck, *supra* note 95, at 323–324.

[193] Article 29 Data Protection Working Party, *supra* note 85, at 21.

[194] *See* Heckmann & Paschke, *supra* note 92, at 379–380.

[195] Franck, *supra* note 149, at 352.

additional assistance to the data subject.[196] This provision was intentionally moved to the non-binding Recitals during trilogue negotiations.[197] Data controllers thus do not have a legal obligation to provide information that will be particularly useful for the data subject to exercise her other rights.

One final comparable restriction is notable concerning Art. 16, the right for the data subject to rectify inaccurate personal data. Data controllers are not required to specify which records most influenced a specific automated decision, which could be extremely helpful to a data subject attempting to identify inaccuracies as grounds to contest a decision. If large amounts of personal data are held, then the subject may have to check tens of thousands of items for inaccuracies.

### 1. Contesting through counterfactuals

Art. 13–15 thus do little to facilitate a data subject's ability to challenge automated decisions. Information is not provided about the safeguards in Art. 22(3) (e.g., the right to contest). It appears that data subjects do not need to be informed of their right not to be subject to an automated decision, which itself could imply a right to contest objectionable automated decisions. Similarly, Recital 71 has neither an explicit link to contesting decisions nor to understanding the black box. Even though an explanation could be helpful, they do not appear to be intended as a precondition for challenging decisions. If explanations were a precondition for contesting decisions, they would appear in the legally binding text.   To offer greater protection to data subjects, these information gaps should be closed, meaning data controllers should inform about the right not to be subject to an automated decision and its safeguards. However, each of these seemingly intentional limitations on the information provided to data subjects suggests that information about the internal logic of an automated decision-making system (in compliance with "meaningful information about the logic involved as well as the significance and the envisaged consequences" in Art. 13(2)(4), 14(2)(g), and 15(1)(h)), which could facilitate contesting decisions, does not need to be provided.[198]

---

[196] Regulation 2016/679, GDPR, recital 60, 2016 O.J. (L 119) 12 (EU).

[197] The European Commission, European Council, and European Parliament in Art. 14(1)(h) proposed to create a legal duty for data controllers to provide any further information beyond those in the notification duties to ensure fair and transparent data processing. *See generally* EUROPEAN COMMISSION, Regulation of the European Parliament and the Council on the Protection of Individuals with Regard to the Processing of Personal Data and on the Free Movement of Such Data (General Data Protection Regulation) (2012), http://ec.europa.eu/justice/data-protection/document/review2012/com_2012_11_en.pdf [https://perma.cc/2HSG-9HX7]; EUROPEAN DIGITAL RIGHTS, *supra* note 186, at 126–127, 129. However, this proposal was not adopted and moved to Recital 60, suggesting that there is no legal duty to provide more information than required in Art 13–14; *see* Franck, *supra* note 99, at 337.

Given these restrictions, counterfactuals could be helpful for contesting decisions, and thus provide greater protection for the data subject than currently envisioned by the GDPR. Regardless of the legal status of the right to explanation, the right to contest is a legally binding safeguard.[199] By providing information about the external factors and key variables that contributed to a specific decision, counterfactuals can provide valuable information for data subjects to exercise their right to contest. This would also be in line with the guidelines of the Article 29 Working Party, which urge that understanding decisions and knowing their legal basis is essential for contesting decisions, and is not necessarily linked to opening the black box.[200] An explanation that low-income led to a loan application being declined could, for example, help the data subject contest the outcome on the grounds of inaccurate or incomplete data regarding her financial situation. Understanding the internal logic of the system that led to income being considered a relevant variable in the decision, which would require a technical explanation unlike a counterfactual explanation (see Appendix 1), may be desirable in its own right, but is not absolutely necessary to contest the decision based on that variable.

Counterfactuals offer a solution and support for contesting decisions by providing data subjects with information about the reasons for a decision, without the need to open the black box. Although Art. 16 of the GDPR gives the data subject the right to correct inaccurate data used to make a decision, the data subject does not need to be informed which data the decision depended. Where a large corpus of data has been collected, an individual without knowledge of which data is relevant or most influential on a particular decision is forced to vet all of it. This lack of information increases the burden on data subjects seeking a different outcome. Counterfactuals provide a compact and easy way to convey these dependencies (i.e., which data was influential), and to facilitate effective claims that a decision was made on the basis of inaccurate data and contest it.

---

[198] Regulation 2016/679, GDPR, art. 15(1)(h), 2016 O.J. (L 119) 43 (EU).
[199] Regulation 2016/679, GDPR, art. 22(3), 2016 O.J. (L 119) 46 (EU).
[200] *See* Article 29 Data Protection Working Party, *supra* note 85, at 25, 27.

### *C. EXPLANATIONS TO ALTER FUTURE DECISIONS*

From the view of the data subject, alongside understanding and contesting decisions, explanations can also be useful to indicate what could be changed to receive a desired result in the future. This purpose does not necessarily relate to the right to contest. Accurate decisions can produce unfavourable results for the data subject. The chances of successfully challenging the decision will also be low in some cases, or the costs and effort required too high. In these situations, the data subject may prefer to change aspects of her situation by adapting her behaviour, and requesting a new decision once more favourable conditions exist.

Using explanations as a guide to altering behaviour to receive a desired automated decision is not directly addressed in the GDPR. This does not, however, undermine the interest data subjects have in receiving desired results from automated decision-making systems. For example, if a subject was rejected for a loan due to insufficient income, a counterfactual explanation will indicate if reapplying in the event of an immediate pay rise is reasonable. The Article 29 Working Party seems to agree, stating in relevant guidelines that "tips on how to improve these habits and consequently how to lower insurance premiums" could be useful for the data subject.[201] For reasons outlined in Appendix 1, technical explanations that try to provide "meaningful information about the logic involved, as well as the significance and the envisaged consequences" of automated decision-making are not guaranteed to be useful in this situation.[202]

Counterfactuals can thus be useful for altering future decisions in favour of the data subject. By providing information about key variables and "close possible worlds" which result in a different decision, data subjects can understand which factors could be changed to receive the desired result. For decision-making models and environments with low variability over time, or models that are "artificially frozen" in time for individuals (i.e., future decisions will be made with the same model as the individual's original decision), this information can help the data subject to alter her behaviour or situation to receive her desired result in the future. Similarly, data controllers could contractually agree to provide the data subject with the preferred outcome if the terms of a given counterfactual were met within a specified period of time.

---

[201] *Id.* at 26.
[202] Regulation 2016/679, GDPR, art. 15, 2016 O.J. (L 119) 43 (EU).

With that said, unanticipated dependencies between intentionally changed attributes and other variables, such as an increase in income resulting from a change in career, may undermine the utility of counterfactuals as guides for future behaviour. Counterfactual explanations can, however, address the impact of changes to more than one variable on a model's output at the same time. Further, regardless of the utility of counterfactuals as guidance for future behaviour, their ability to help individuals understand which data and variables were influential in specific prior decisions remains unaffected.

## CONCLUSION

We have proposed a novel lightweight form of explanation that we refer to as counterfactual explanations. Unlike existing approaches that try to provide insight into the internal logic of black box algorithms, counterfactual explanations do not attempt to clarify how decisions are made internally. Instead, they provide insight into which external facts could be different in order to arrive at a desired outcome.[203] Importantly, counterfactual explanations are efficiently computable for many standard classifiers, particularly neural networks. As our new form of explanation significantly differs from existing works, we have justified its nature as an explanation with reference to previous works in the philosophical literature and early A.I.

From the view of the data subject, we have assessed three purposes of explanations of automated decisions: understanding, contesting and altering. We compared these aims with the provisions of the GDPR and evaluated if they rely upon opening the black box. We concluded that the framework offers little support to achieve these goals, and does not mandate that algorithms are explainable to understand, contest or alter decisions.

The GDPR itself provides little insight into the intended purpose and content of explanations. Recital 71, the only provision that explicitly mentions explanations, does not reveal their intended purpose or content. Given the final text of the GDPR, it appears that explanations can voluntarily be offered after decisions have been made, and are not a required precondition to contest decisions. Further, there is no clear link that suggests that explanations under Recital 71 require opening the black box.

---

[203] The method described here is compatible with a proposal made by Citron and Pasquale to allow consumers to manually enter "hypothetical alterations" to their credit histories and view their effects. Danielle K. Citron & Frank A. Pasquale, The Scored Society: Due Process for Automated Predictions, 89 Wash. L. Rev. 1, 28 (Jan. 2014) https://papers.ssrn.com/abstract=2376209 [https://perma.cc/VQA6-2MBT].

Recognising this relative lack of insight into explanations, related provisions addressing automated decision-making were examined. Notification duties defined in Art. 13–14 apply prior to data processing or before a decision is made (i.e. at the time of data collection), and provide a simple and generic overview of intended data processing activities that aims to inform a general audience.[204] This type of "meaningful overview" of automated decision-making is largely unsuitable to understand the rationale of specific decisions. Art. 13–14 similarly do not facilitate contesting decisions, owing to a lack of information to be provided about the right not to be subject to automated individual decision-making,[205] and its safeguards.[206] The right of access[207] provides nearly identical information to Art. 13–14, and thus offers similarly limited value for understanding and contesting decisions. The rights and freedoms of others (e.g. privacy or trade secrets) which are protected in Art. 15(4) and Recital 63 pose an additional barrier to transparency when access requests are lodged. Across each of these Articles, technical explanations of the internal logic of automated decision-making systems are not legally mandated. Finally, offering explanations to give guidance how to receive the desired result in the future does not appear to be an aim of the GDPR, but could still be highly useful for individuals seeking alternative, more desirable outcomes.

Any future attempt to implement a legally binding right to explanation as a safeguard against automated decision-making within the framework provided by the GDPR faces several notable challenges. Automated decision-making must be based "solely on automated processing," and have "legal effects" or similarly significant effects.[208] Additionally, exemptions from the safeguards against automated decision-making can be introduced through Member State law.[209]

However, the data subject's desire to understand, contest, and alter decisions does not change based on these definitional issues. We therefore propose to move past the limitation of the GDPR and to use counterfactuals as unconditional explanations. These unconditional explanations should be given whenever requested, regardless of outcome (positive or negative decision), whether the decision was based on solely automated processes and their (legal or similar significant) effects.

[204] Regulation 2016/679, GDPR, art. 12(7), 2016 O.J. (L 119) 40 (EU).
[205] Regulation 2016/679, GDPR, art. 22(1), 2016 O.J. (L 119) 46 (EU).
[206] Regulation 2016/679, GDPR, art. 22(3), 2016 O.J. (L 119) 46 (EU).
[207] Regulation 2016/679, GDPR, art. 15, 2016 O.J. (L 119) 43 (EU).
[208] Regulation 2016/679, GDPR, art. 22, 2016 O.J. (L 119) 46 (EU).
[209] Regulation 2016/679, GDPR, art. 23, 2016 O.J. (L 119) 46–47 (EU). These exemptions may be based, for example, on national security; the enforcement of civil law claims; or the protection of the data subject or the rights and freedoms of others.

Counterfactual explanations could be implemented in several ways. The transience of decision-making models suggests that counterfactuals either need to be computed automatically at the time a decision is made, or a copy of the model archived to compute counterfactuals at a later time. As multiple outcomes based on changes to multiple variables may be possible, a diverse set of counterfactual explanations should be provided, corresponding to different choices of nearby possible worlds for which the counterfactual holds. These sets could be disclosed when automated decision-making occurs, or in response to specific requests lodged by individuals or a trusted third party auditor.[210] In any case, disclosures should occur in a reasonable window of time.[211]

Future research should determine appropriate distance metrics and requirements for a sufficient and relevant set of counterfactuals across use sectors and cases which have very different needs. While prior philosophical debate may prove helpful, the absence of causal models in most modern classifiers, as well as the preferences of the recipient(s) of the set, must be accounted for in choosing appropriate metrics and requirements. Compared to prior discussion of measuring "closest possible worlds," setting requirements for appropriate "close possible worlds" represents a very different philosophical, social, and legal challenge.

To minimise bureaucratic burdens for data controllers and delays for data subjects and third party auditors, automated calculation and disclosure of counterfactuals would be preferable. We recommend this type of automated implementation going forward. One possible approach is to provide individuals or third party auditors with access to "auditing APIs,"[212] which allow users to request counterfactual explanations from the service provider, and perhaps compute them directly via the API. Access to historical decision-making models used for the decision at hand, as well as permissive terms of service that allow for such auditing, would be required.[213] This functionality could potentially be embedded in existing APIs.

---

[210] *See generally* Wachter, Mittelstadt & Floridi, *supra* note 1.

[211] *See, e.g.*, Regulation 2016/679, GDPR, art. 12(3), 2016 O.J. (L 119) 40 (EU).

[212] Who would shoulder the costs of hosting these APIs and computing counterfactuals is an important political issue that would require resolution. This issue goes beyond the scope of this paper. For related discussion of implementing algorithmic auditing, see Christian Sandvig et al., *Auditing Algorithms: Research Methods for Detecting Discrimination on Internet Platforms*, DATA & DISCRIMINATION: CONVERTING CRITICAL CONCERNS INTO PRODUCTIVE INQUIRY (May 22, 2014),

Counterfactual explanations provide reasons why a particular decision was received (e.g., low income), offer grounds to contest it (e.g., if the data controller used inaccurate data about the income of the applicant), and provide limited "advice" on how to receive the desired results in the future (e.g., an increase of 4000 pounds/year would have resulted in a positive application). Their usage would help to resolve two primary objections to a legally binding right to explanation: first, that explaining the internal logic of automated systems to experts and non-experts alike is a highly difficult and perhaps intractable challenge; and second, that an excessive disclosure of information about the internal logic of a system could infringe on the rights of others, either by revealing protected trade secrets or by violating the privacy of individuals whose data is contained in the training dataset. In contrast, counterfactuals allow an individual to receive explanations without conveying the internal logic of the algorithmic black box (beyond a limited set of dependencies), and are less likely to infringe the rights and freedoms of others than full disclosure. Assuming reasonable limitations are set on the number of counterfactuals that must be provided, counterfactuals are also less likely to provide information that reveals trade secrets or allows gaming of decision-making systems.

As a minimal form of explanation, counterfactuals are not appropriate in all scenarios. In particular, where it is important to understand system functionality, or the rationale of an automated decision, counterfactuals may be insufficient in themselves. Further, counterfactuals do not provide the statistical evidence needed to assess algorithms for fairness or racial bias. Given these limitations, more general forms of explanations and interpretability should still be pursued

---

http://social.cs.uiuc.edu/papers/pdfs/ICA2014-Sandvig.pdf [https://perma.cc/9NKH-J69E]; Brent Mittelstadt, *Auditing for Transparency in Content Personalization Systems*, 10 Int'l. J. Comm. 12 (2016) http://ijoc.org/index.php/ijoc/article/view/6267 [https://perma.cc/TCN4-56QU].

[213] Counterfactuals must be computed on the basis of the decision-making model at the time the decision was taken. Assuming automated decision-making models change over time, in implementations not involving automatic computation of counterfactuals at the time a decision is made (which may be cost prohibitive), it will be necessary for data controllers to keep 'audit logs' indicating the state of the decision-making model at the time of the decision.

to increase accountability and better validate the fairness and functionality of systems.

However, counterfactuals represent an easy first step that balances transparency, explainability, and accountability with other interests such as minimising the regulatory burden on business interest or preserving the privacy of others, while potentially increasing public acceptance of automatic decisions. Rather than waiting years for jurisprudence to dissolve all these uncertainties, we propose to abandon the narrow definitions and conditions the GDPR imposes on automated decision-making, and offer counterfactuals as unconditional explanations at the request of affected individuals.

APPENDIX 1: SIMPLE LOCAL MODELS AS EXPLANATIONS

As discussed in Section II.B, 'Explanations in A.I. and Machine Learning,' approaches such as LIME[214] that generate simple models as local approximations of decisions make a three-way trade-off between the quality of the approximation versus the ease of understanding the function and the size of the domain for which the approximation is valid.[215]

To illustrate the instabilities of the approach with respect to the size of the domain, we consider a simple function of one variable. Even for problems such as this, the notion of scale, or how large a region should an explanation try to describe, is challenging with the ideal choice of scale depending on what the explanation would be used for.

As a real-world example, consider being stopped by someone in a car who asks which direction they should travel in to go north. Fundamentally, this is a difficult question to answer well, with the most appropriate answer depending upon how far north they wish to travel. If they do not intend to travel far, simply pointing north gives them enough information. However, if they intend to travel further, roads that initially point north may double back on themselves or be cul-de-sacs and better directions are needed. If they intend to travel a long way, they may be better off ignoring the compass bearing entirely, and instead try to directly join up with an inter-city network.

This exact issue is faced when automating explanations of decisions: the generated explanations are generic, and designed to be useful to the recipient of the explanation regardless of how they are used. However, as shown in Figure 1, the explanation — or simplified model — can vary wildly with the scale or range of inputs considered.

---

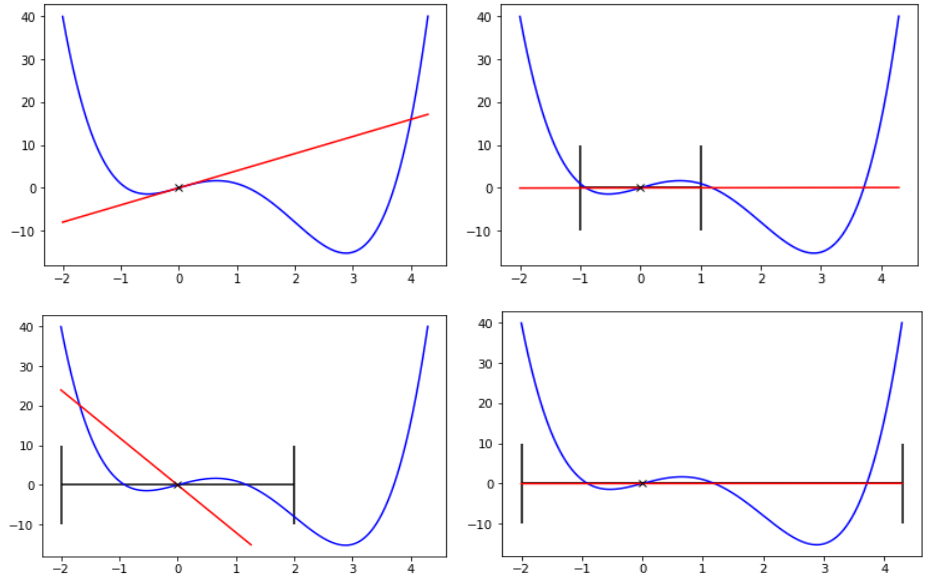

*Figure 1 - **Local models varying with choice of scale.** The red line in each subfigure shows a local approximation of the same blue score curve centred at the same location in each plot. The varying range over which the approximation is computed is given by the region marked by black bars. Different choices of range e.g. top left vs. bottom left can lead to completely opposing explanations where the score either increases or decreases as the value along the bottom axis increase.*

As can be seen, the direction and magnitude of the linear approximation (red) to underlying function (blue) vary dramatically with choice of domain, and deciding which approximation is most helpful to a layperson trying to understand the decision made about them is non-trivial.

To show the difficulties that would exist in either trying to use local models to either compute counterfactuals, or simply for the data subject to adjust their score, we assume that the subject desired to know how to obtain a lower score of -10 or below. In this case, none of the local approximations would be useful. The top left model, which is based on exact description of the function around point x predicts that a score of -10 would be obtained with a value of -2.5—corresponding to an actual score of 91.5, while the two centre approximations suggest that it is not possible to obtain any score except 0, and the bottom left approximation says that -10 occurs at near 0.9—which actually corresponds to a local maxima.

In contrast, the counterfactual explanation for a query such as "Why was the score not below -10?" would return the answer "Because the x value was not 2.15" (the counterfactual is illustrated in figure 2 by the green dot). Of course, it should be noted that the two approaches are generally incomparable. In much the same way, if a data subject desired to know a local linear approximation about their data point, knowledge of counterfactuals would not be helpful. However, of the two approaches, counterfactuals are the only one that will provide some indication if it is worth reapplying for a loan in the event of a pay rise.

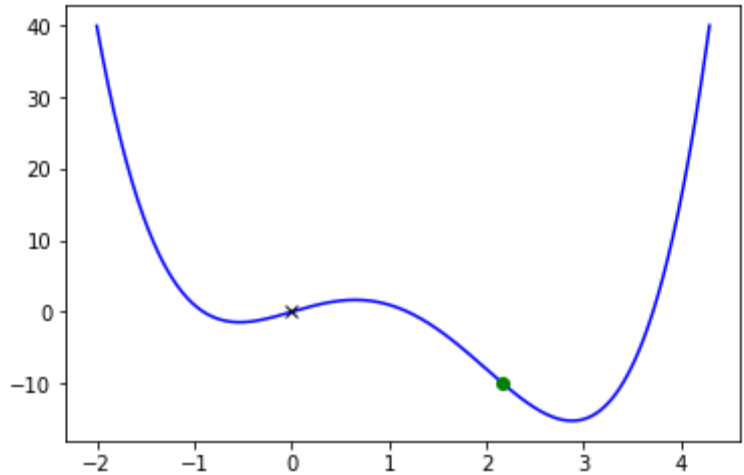

***Figure 2 – Visual representation of the range of a counterfactual explanation***

APPENDIX 2: EXAMPLE TRANSPARENCY INFOGRAPHIC

The figure below shows several icons proposed by the European Parliament during trilogue that were ultimately not adopted as a standard. They nonetheless reveal the level of complexity expected by EU legislators when communicating information to data subjects under Art. 13–14. The reliance on generic icons suggests that individual-level, contextualised information is not required, meaning Art. 13–14 are not intended to provide a 'de facto' right to explanation comparable to the right contained in Recital 71. The relative simplicity of the icons also suggests that a broad audience is intended, comparable for example to website privacy notices. Finally, although the icons were rejected, the EC has been tasked with developing such standardised icons in the future (Art. 12(8)), meaning comparable icons are seen as an acceptable way to convey the information required by Art. 13–14.

| ICON | ESSENTIAL INFORMATION | FULFILLED |
|------|----------------------|-----------|
| 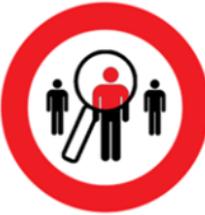 | No personal data are **collected** beyond the minimum necessary for each specific purpose of the processing | |
| 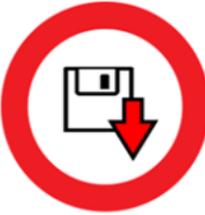 | No personal data are **retained** beyond the minimum necessary for each specific purpose of the processing | |
| 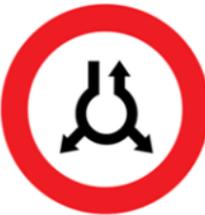 | No personal data are **processed** for purposes other than the purposes for which they were collected | |
| 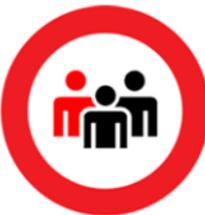 | No personal data are **disseminated** to commercial third parties | |
| 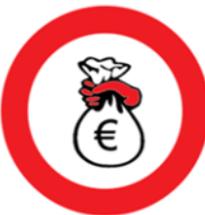 | No personal data are **sold or rented out** | |
| 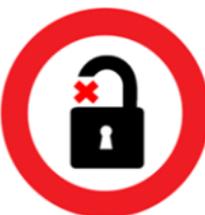 | No personal data are retained in **unencrypted** form | |

COMPLIANCE WITH ROWS 1-3 IS REQUIRED BY EU LAW